\documentclass[sn-mathphys-num,iicol]{sn-jnl}
\usepackage{graphicx}%
\usepackage{multirow}%
\usepackage{amsmath,amssymb,amsfonts}%
\usepackage{amsthm}%
\usepackage{mathrsfs}%
\usepackage[title]{appendix}%
\usepackage{manyfoot}%
\usepackage{booktabs}%
\usepackage{algorithmicx}%
\usepackage{algpseudocode}%
\usepackage{listings}%
\usepackage[linesnumbered,ruled]{algorithm2e}
\usepackage{bbm}
\usepackage{bbding}
\usepackage{pifont}
\usepackage{threeparttable}
\usepackage{ragged2e}
\usepackage{colortbl}
\usepackage{float}
\usepackage{comment}
\usepackage{makecell}
\usepackage{titletoc}
\usepackage{setspace}
\usepackage[table]{xcolor}
\usepackage{epsfig}
\usepackage{newtxtext,newtxmath}
\usepackage{tabularx}

\usepackage{caption}
\theoremstyle{thmstyleone}%
\theoremstyle{thmstyletwo}%

\theoremstyle{thmstylethree}%

\begin{document}
\title[DynamicVis]{DynamicVis: Dynamic Visual Perception for Efficient Remote Sensing Foundation Models}

\author[1,2]{\fnm{Keyan} \sur{Chen}}

\author[1]{\fnm{Chenyang} \sur{Liu}}

\author[1]{\fnm{Bowen} \sur{Chen}}

\author[3]{\fnm{Wenyuan} \sur{Li}}

\author[1]{\fnm{Zhengxia} \sur{Zou}}

\author[2]{\fnm{Shijian} \sur{Lu}}

\author*[1,]{\fnm{Zhenwei} \sur{Shi}}\email{shizhenwei@buaa.edu.cn}

\affil*[1]{\orgname{Beihang University}, \orgaddress{\state{Beijing}, \country{China}}}

\affil[2]{\orgname{Nanyang Technological University}, \orgaddress{\country{Singapore}}}

\affil[3]{\orgname{University of Hong Kong}, \orgaddress{\state{Hong Kong}, \country{China}}}

\abstract{
  The continuous advancement of remote sensing technology has enabled high-resolution Earth observation; however, interpreting these expansive images using modern Vision Foundation Models (VFMs) remains a significant challenge. Unlike object-centric natural images, remote sensing imagery is fundamentally characterized by extreme target sparsity and massive spatial redundancy. Key objects of interest (e.g., ships, vehicles, scattered buildings) often occupy less than 1\% of the spatial extent, surrounded by vast, target-free backgrounds. Existing VFMs predominantly rely on uniform dense processing (e.g., Vision Transformers) and pixel-reconstruction pre-training paradigms (e.g., Masked Autoencoders). These approaches inherently waste substantial computational capacity on modeling redundant backgrounds and inadvertently dilute the feature representations of small, sparse targets. To bridge this structural misalignment, we propose \textbf{\textit{DynamicVis}}, a visual foundation model explicitly tailored to the sparse nature of remote sensing imagery. Architecturally, DynamicVis introduces a Dynamic Region-Aware State Space Model (SSM) that bypasses uniform computation. It adaptively routes and incrementally models only task-relevant, high-salience tokens while employing a parameter-free integration for background context, drastically reducing the complexity of processing ultra-long 2D token sequences ($\sim$100,000). Crucially, to equip the network with robust spatial-selection capabilities, we propose a novel Region-Level Meta-Embedding Multi-Instance Learning (MIL) pre-training paradigm. Trained on a million-scale dataset, this paradigm explicitly disentangles sparse foreground instances from dense backgrounds in the latent semantic space, overcoming the semantic ambiguity of conventional pixel-reconstruction methods. Extensive evaluations across nine diverse downstream tasks reveal that DynamicVis exhibits exceptional efficacy, particularly dominating in sparse-target and instance-level perception tasks (e.g., small object detection, instance segmentation, and change detection). Notably, DynamicVis processes high-resolution imagery (2048 {\unboldmath$\times$} 2048) with merely 97 ms latency and 833 MB of memory consumption without employing any acceleration techniques (approximately 6\% and 3\% of a base ViT, respectively), establishing an optimal trade-off between computational scalability and representation accuracy for multi-granular remote sensing analysis. Source code is available at \url{https://github.com/KyanChen/DynamicVis}.
}




\maketitle

\section{Introduction}\label{sec:intro}

The rapid evolution of Earth observation technologies has ushered in an era of ultra-high-resolution remote sensing (RS), providing unprecedented granular details for critical applications such as urban planning, maritime surveillance, and disaster management \cite{sun2022ringmo, cong2022satmae, reed2023scale, zheng2024changen2,xiong2024neural, liu2024remoteclip, kuckreja2024geochat}. To efficiently interpret these massive volumes of multimodal data, Vision Foundation Models (VFMs) pre-trained on large-scale datasets have emerged as the dominant paradigm \cite{klemmer2025satclip, zhang2024rs5m, wang2022advancing, xiao2025foundation}. By learning transferable and generalizable visual representations, these models aim to satisfy the heterogeneous demands of diverse downstream tasks with minimal task-specific adaptation costs \cite{jakubik2023foundation, bastani2023satlaspretrain, guo2024skysense, chen2023continuous,fuller2023croma, manas2021seasonal}.

Despite this promising progress, adapting general-purpose foundation models to the RS domain encounters a fundamental structural bottleneck rooted in the unique physical properties of satellite imagery: extreme target sparsity and massive spatial redundancy \cite{smith2023earthpt, lacoste2023geo, lu2025vision, wang2023samrs,zhu2026foundations}. In natural object-centric images (e.g., ImageNet), targets typically occupy the center and constitute the majority of visual content. Conversely, RS images cover expansive geographical areas where key objects of interest (e.g., small vehicles, maritime vessels, scattered buildings) often occupy less than 1\% of the spatial extent. These targets are typically submerged within extensive backgrounds such as oceans, deserts, or persistent cloud cover \cite{wang2024skyscript, weng2025vision, wu2025semantic, zou2023object}.

Existing RS foundation models (e.g., RingMo \cite{sun2022ringmo, yao2023ringmo}, SpectralGPT \cite{hong2024spectralgpt}) generally inherit Vision Transformer (ViT) architectures and rely heavily on Masked Autoencoder (MAE) pre-training paradigms \cite{he2022masked,dosovitskiy2020image}. We argue that this reliance creates a twofold inductive misalignment. First, the uniform quadratic self-attention mechanism in ViTs allocates equal computational resources to dense backgrounds and sparse foregrounds \cite{dosovitskiy2020image, rao2021dynamicvit, bolya2022token}. For high-resolution RS imagery ($\sim$ millions of pixels), this renders the processing of redundant background tokens computationally prohibitive \cite{wang2022advancing,jiang2024lemevit,guo2024skysense}. Second, pixel-level reconstruction pre-training (exemplified by MAE) forces the network capacity to over-index on memorizing repetitive background textures rather than learning highly discriminative features for sparse, semantically rich targets \cite{he2022masked, cong2022satmae, sun2022ringmo}.

\begin{figure*}[t]
  \centering
  \includegraphics[width=\linewidth]{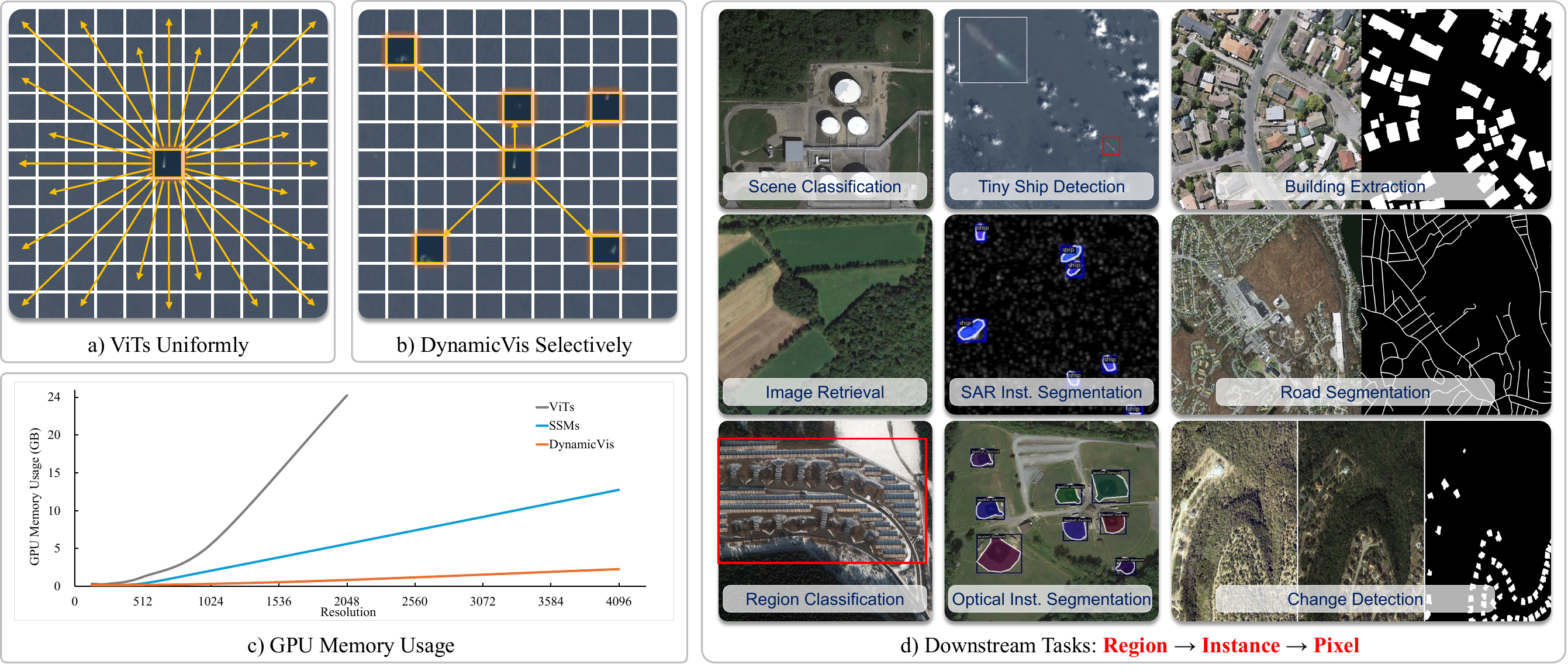}
\caption{Motivation and Overview of DynamicVis. \textbf{a)} ViTs process all visual tokens uniformly, leading to massive computational waste on redundant remote sensing backgrounds. \textbf{b)}: DynamicVis introduces an adaptive token routing mechanism combined with SSMs to selectively process only high-salience regions, explicitly addressing the spatial sparsity of key targets. \textbf{c)}: Memory consumption comparison, demonstrating the superior computational scalability of DynamicVis at ultra-high input resolutions. \textbf{d)}: Comprehensive evaluations across nine downstream tasks demonstrate its efficacy, particularly in sparse-target perception.} \label{fig:vis_teaser}
\vspace{-0.3cm}
\end{figure*}

To bridge this gap, we rethink the design of RS foundation models by introducing ``sparse spatial perception" as a core inductive bias (see Fig. \ref{fig:vis_teaser}). Motivated by the selective attention mechanisms of the biological visual system, we propose \textbf{\textit{DynamicVis}}, an efficient foundation model explicitly designed to navigate the spatial redundancy of Earth observation data. Architecturally, we construct a Dynamic Region-Aware State Space Model (SSM) backbone. While recent SSMs (e.g., Mamba) offer linear computational scaling for long sequences, blindly applying them to raw RS images still mandates the processing of massive amounts of redundant background tokens \cite{gu2024mamba, zhu2024vision, liu2024vmamba, chen2024rsmamba}. Instead, DynamicVis employs an adaptive token routing mechanism across its hierarchical layers. It dynamically identifies a sparse subset of highly salient tokens via learnable importance scores, performs incremental modeling exclusively on these critical regions using dual-path SSM scans, and utilizes parameter-free integration for the background context. This selective routing drastically reduces computational overhead while preserving the fine-grained structural integrity vital for small-object analysis.


Crucially, equipping a network with the ability to dynamically ``drop" sparse-information tokens necessitates a pre-training paradigm that explicitly teaches the model how to distinguish salient semantics from background noise. Conventional image-level contrastive learning (e.g., CLIP) provides supervisory signals that are too coarse to isolate small targets, while MAE reconstruction lacks the high-level semantic alignment required for logical routing \cite{radford2021learning, zhong2022regionclip, he2022masked}. To address this, we propose a novel Region-Level Meta-Embedding Multi-Instance Learning (MIL) pre-training paradigm. Leveraging a million-scale dataset with weak region-level annotations (fMoW \cite{christie2018functional}), this paradigm treats each massive image as a ``bag" of instances. By conducting contrastive learning between regional visual embeddings and categorical meta-embeddings in the latent space, the model is compelled to explicitly decouple heterogeneous foreground instances from expansive backgrounds. This mechanism provides the precise semantic guidance required for the dynamic routing module to function optimally during downstream fine-tuning.

Extensive evaluations demonstrate that DynamicVis achieves state-of-the-art performance across nine diverse interpretation tasks. Acknowledging its strong inductive bias toward spatial sparsity, DynamicVis maintains competitive performance on dense continuous prediction tasks while exhibiting absolute superiority in tasks requiring sparse and discrete target perception, such as tiny object detection, instance segmentation, and bi-temporal change detection.

The principal contributions of this work are summarized as follows:




i) We rethink remote sensing foundation models through the perspective of spatial sparsity, proposing DynamicVis. By embedding a dynamic visual perception mechanism, the model effectively overcomes the computational barriers of high-resolution inputs while maximizing feature extraction for sparse targets.

ii) We introduce a Dynamic Region-Aware SSM architecture. By coupling adaptive token routing with linear state-space models, we achieve a highly scalable encoding that mitigates the background redundancy prevalent in traditional dense processing frameworks.

iii) We design a Meta-Embedding MIL pre-training paradigm utilizing large-scale weak region annotations. This strategy bridges the gap between semantic clustering and spatial sparsity, providing robust guidance to identify salient regions dynamically and avoiding the texture-memorization pitfalls of MAE.

iv) DynamicVis achieves exceptional efficiency and robust generalization. Processing a high-resolution image ($2048 \times 2048$) requires only 97 ms latency and 833 MB of GPU memory (approximately 6\% and 3\% of a ViT-base, respectively) without employing acceleration techniques. It establishes an optimal trade-off between computational scalability and representation accuracy for multi-granular RS analysis.

\section{Related Works}

\subsection{Representation Learning and Vision Foundation Models in Remote Sensing}

Representation learning in the geospatial domain has undergone a fundamental paradigm shift, transitioning from training task-specific architectures from scratch to fine-tuning large-scale VFMs \cite{li2026agrifm, li2021geographical, sun2022ringmo, hong2024spectralgpt}. Given the chronic scarcity of high-quality, pixel-level annotations in remote sensing, unsupervised and self-supervised learning have naturally emerged as the dominant pre-training strategies \cite{wang2022self, tao2023self}. Currently, Masked Image Modeling (MIM), exemplified by the MAE framework, is the most prevalent approach. Pioneering models such as RingMo \cite{sun2022ringmo}, SatMAE \cite{cong2022satmae}, and SpectralGPT \cite{hong2024spectralgpt} have demonstrated that reconstructing masked patches can yield robust multi-spectral and multi-temporal representations.

Despite their success in capturing generic textural patterns, MAE-based models suffer from a critical inductive misalignment with the inherent physical properties of Earth observation imagery. Unlike natural images, satellite scenes are fundamentally characterized by massive backgrounds (e.g., expansive oceans, bare land, or persistent cloud cover) \cite{jiang2024lemevit, reed2023scale}. Consequently, pixel-level reconstruction inadvertently forces the network's capacity to over-index on memorizing these redundant background statistics, rather than learning highly discriminative features for sparse, semantically rich targets. Conversely, contrastive learning approaches, such as SeCo \cite{manas2021seasonal} and SkySense \cite{guo2024skysense}, focus on aligning global image-level representations. Unfortunately, this coarse semantic alignment tends to dilute the localized, fine-grained details strictly required for small object \cite{wang2021dense}. Meanwhile, recent prompt-based models like RSPrompter achieve excellent instance-level discrimination but rely heavily on frozen weights from the Segment Anything Model (SAM), restricting their generalizability across non-segmentation downstream tasks \cite{chen2024rsprompter, kirillov2023segment}.

To bridge this semantic gap, DynamicVis introduces a novel Region-Level Meta-Embedding Multi-Instance Learning (MIL) pre-training paradigm. The core motivation stems from the prerequisite of our dynamic spatial perception mechanism: to accurately route salient tokens, the model must fundamentally understand how to distinguish discrete foreground instances from dense backgrounds in the latent space. Rather than reconstructing raw pixels or performing coarse global contrastive learning, our MIL approach explicitly treats high-resolution scenes as ``bags" of instances. By aligning regional visual embeddings with categorical meta-embeddings, the model is compelled to decouple heterogeneous foreground targets from expansive backgrounds. This pre-training mechanism inherently equips the network with the precise semantic guidance necessary to power the dynamic routing module during downstream adaptation.

\subsection{Efficient Architectures and Vision State Space Models}

The quadratic computational complexity of the self-attention mechanism in ViTs presents a prohibitive bottleneck for processing high-resolution remote sensing images \cite{dosovitskiy2020image}. To mitigate this computational burden, researchers have extensively explored hierarchical architectures (e.g., Swin Transformer \cite{liu2021swin}) and modern efficient CNNs (e.g., ConvNeXt \cite{liu2022convnet}). Recently, linear attention mechanisms, particularly Selective State Space Models (SSMs) like Mamba \cite{gu2024mamba}, have garnered significant attention. By parameterizing input-dependent state transitions, SSMs successfully achieve linear computational complexity relative to sequence length. This architectural breakthrough has been rapidly adapted for vision tasks via models like Vim \cite{zhu2024vision} and VMamba \cite{liu2024vmamba}, and subsequently introduced to the remote sensing domain through task-specific adaptations such as RSMamba \cite{chen2024rsmamba}, CDMamba \cite{zhang2025cdmamba}, and ChangeMamba \cite{chen2024changemamba}.

However, while existing Mamba-based vision models successfully reduce complexity from $\mathcal{O}(N^2)$ to $\mathcal{O}(N)$, they strictly adhere to a dense processing paradigm. This means they still allocate computational resources uniformly across all visual tokens \cite{rao2021dynamicvit, yang2025smamba}. For a typical $2048 \times 2048$ remote sensing image, this mandates the processing of hundreds of thousands of redundant background tokens that contribute minimally to the final interpretation \cite{zhang2023efficient}. In essence, simply replacing ViTs with SSMs merely alleviates the sequence-length bottleneck without addressing the fundamental issue of massive spatial redundancy.

DynamicVis diverges fundamentally from existing SSM adaptations by embedding an Adaptive Token Routing mechanism directly into the SSM backbone. Rather than densely scanning the entire image, DynamicVis dynamically activates and routes only a sparse subset of high-salience tokens through computationally intensive dual-path SSM scans. By integrating sparsity directly into the linear modeling process, DynamicVis maximizes the utility of SSMs, achieving an ultra-scalable encoding process explicitly tailored to the sparse nature of Earth observation data.

\subsection{Dynamic Token Sparsification and Structural Preservation}

The concept of dynamically selecting salient tokens to accelerate vision models has been explored in natural image and video domains. Pioneering works such as DynamicViT \cite{rao2021dynamicvit} utilize prediction modules to progressively discard uninformative tokens. TokenLearner \cite{ryoo2021tokenlearner} learns to aggregate spatial-temporal concepts into a handful of tokens, while EVT \cite{wang2022efficient} introduces spatial-temporal token selection for video transformers. Other token aggregation techniques rely on k-NN clustering (e.g., NFormer \cite{wang2022nformer}) or graph-based grouping (e.g., LLaVA-PruMerge \cite{shang2025llava}).

Although these token-reduction strategies are highly effective for object-centric natural images where the background can often be safely discarded, directly applying them to remote sensing introduces severe inductive biases. In Earth observation imagery, key objects of interest (e.g., a $10 \times 10$ pixel ship in a $1024 \times 1024$ image) are extremely small and sparsely distributed \cite{ding2021object, xie2021oriented}. Aggressive token pruning strategies, guided merely by global classification logits, often permanently discard these critical small targets. Furthermore, completely dropping spatial tokens irrevocably severs the continuous structural integrity required for dense prediction tasks \cite{bolya2022token,chen2022vision}. This inevitably leads to catastrophic performance drops in pixel-level tasks when compared to specialized segmentation models like PoolFormer \cite{yu2022metaformer}, RoadFormer \cite{jiang2022roadformer}, or SegFormer \cite{xie2021segformer}.

To resolve the inherent tension between token reduction and structural preservation, DynamicVis introduces a Region-Aware Incremental Modeling strategy. Instead of permanently discarding unselected background tokens, our dynamic router explicitly identifies sparse foreground tokens for deep contextual refinement via SSMs. Concurrently, the unselected background tokens bypass the heavy computation and are reintegrated into the feature map via parameter-free residual connections. This design guarantees significant computational efficiency without sacrificing the contextual awareness and spatial continuum essential for precise, pixel-level remote sensing interpretation.

\section{Methodology}

\subsection{Overview}
Earth observation imagery is fundamentally characterized by massive spatial redundancy and the extreme sparsity of targets of interest \cite{guo2024skysense,sun2022ringmo}. Motivated by these geometric properties, we propose \textbf{\textit{DynamicVis}}, an efficient visual foundation model explicitly engineered to decouple sparse foreground instances from complex backgrounds. As illustrated in the unified pipeline, DynamicVis shifts away from the uniform dense processing paradigm characteristic of ViTs \cite{bolya2022token, rao2021dynamicvit}. Instead, it introduces a Dynamic Region-Aware SSM Backbone. This architecture adaptively routes high-salience, task-relevant tokens for computationally intensive sequence modeling, while integrating background context via a parameter-free mechanism to preserve the macroscopic spatial topology.

Furthermore, to provide the dynamic routing module with robust semantic guidance for distinguishing salient foregrounds from background noise, we propose a Region-Level Meta-Embedding MIL pre-training paradigm. Operating on a million-scale dataset with weak region-level annotations, this paradigm aligns regional visual representations with categorical meta-embeddings in a continuous latent space. This approach effectively circumvents the texture-memorization pitfalls inherent to pixel-reconstruction methods like MAE \cite{he2022masked}. To adapt the foundation model to diverse downstream applications, spanning region-, instance-, and pixel-level tasks, the architecture integrates task-specific decoders. The unified interpretation workflow is formalized as:
\begin{equation}
\mathcal{R} = \Phi_{\mathrm{decoder}}^{\Gamma} \left( \Phi_{\mathrm{backbone}}(\mathcal{I}) \right),
\end{equation}
where $\mathcal{I}$ denotes the input high-resolution image, $\Phi_{\mathrm{backbone}}$ represents the shared dynamic foundation backbone, $\Phi_{\mathrm{decoder}}^{\Gamma}$ indicates the parameterized decoder tailored for a specific downstream task type $\Gamma$, and $\mathcal{R}$ encompasses the final interpretation outputs (e.g., categorical labels, object bounding boxes, or dense pixel-wise masks).

\begin{figure*}[!htbp]
\centering
\includegraphics[width=\linewidth]{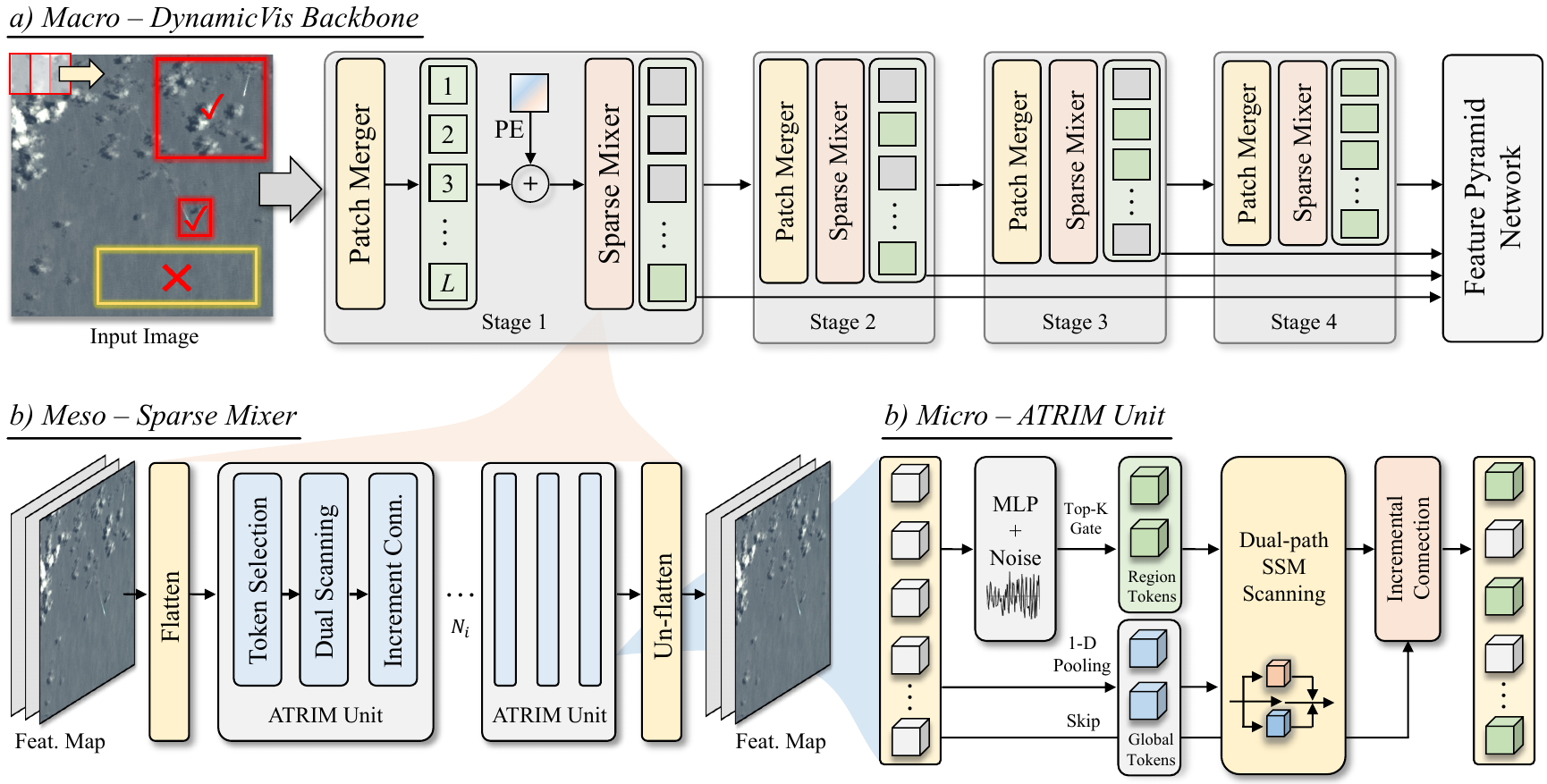}
\caption{
\textbf{a)} Overview of the Dynamic Region-Aware SSM Backbone. The architecture leverages a progressive, small-stride patch merger to preserve tiny-target details and employs dynamic sparse mixers across four hierarchical stages. By actively routing only high-salience tokens (red boxes) into the SSM while bypassing repetitive background patches (yellow boxes), the backbone efficiently generates multiscale semantic representations without destroying spatial topology. \textbf{b)} Structural breakdown of the Dynamic Sparse Mixer. It comprises a flattening operation, $N_i$ Adaptive Token Routing and Incremental Modeling (ATRIM) units designed to selectively aggregate global semantics, and an unflattening operation to restore 2D spatial arrangements for various downstream tasks. \textbf{c)} Detailed architecture of the ATRIM unit. Instead of permanently discarding tokens, the module routes highly salient tokens for heavy SSM refinement, while unselected background tokens bypass the intense computation and are restored via a context-preserving residual connection.
} \label{fig:vis_model_overview}
\vspace{-0.3cm}
\end{figure*}

\subsection{Dynamic Region-Aware SSM Backbone}

\subsubsection{Overall Architecture}
The backbone constitutes the perception core of DynamicVis, specifically designed to balance fine-grained local detail extraction with highly efficient global context integration. As illustrated in Fig. \ref{fig:vis_model_overview} a), the architecture comprises a multi-scale dynamic feature extractor coupled with a Feature Pyramid Network (FPN) \cite{lin2017feature}. The hierarchical feature extraction process is expressed as:
\begin{equation}
\{\mathbf{F}_i\}_{i=1}^5 = \Phi_{\mathrm{backbone}}(\mathcal{I}) = \Phi_{\mathrm{fpn}} \circ \Phi_{\mathrm{ms\text{-}extract}}(\mathcal{I}).
\end{equation}
Given an input image $\mathcal{I} \in \mathbb{R}^{H \times W \times 3}$, the multi-scale extractor $\Phi_{\mathrm{ms\text{-}extract}}$ processes the visual signal through four hierarchical stages. To address the severe scale variations typical of remote sensing targets, the FPN ($\Phi_{\mathrm{fpn}}$) subsequently performs top-down cross-scale feature fusion. This mechanism reconstructs multi-scale feature maps $\{\mathbf{F}_i\}$ at five distinct resolution levels ($i \in \{1,2,3,4,5\}$) with a uniform channel dimension $d$ (e.g., $d=256$), ensuring seamless compatibility with standard dense prediction heads (e.g., Mask R-CNN \cite{he2017mask}, UperNet \cite{xiao2018unified}).

\subsubsection{Spatial-Preserving Hierarchical Patch Merger}

Unlike ViTs that employ an aggressive $16 \times 16$ patch embedding, which irreversibly compromises fine-grained spatial configurations essential for resolving tiny targets (e.g., small vehicles or sparse ships), DynamicVis adopts a progressive, small-stride downsampling strategy. At the $i$-th stage, the spatial-preserving patch merger utilizes a 2D convolutional block with a kernel size of $k_i \times k_i$ and a small stride of $s_i \times s_i$ (typically $s_i=2$):
\begin{equation}
\begin{aligned}
\mathbf{F}_i &= \Phi_{\mathrm{Conv2D}}(\mathbf{F}_{i-1}; k_i, s_i), \\
&\text{where } \mathbf{F}_i \leftarrow \mathbf{F}_i + \mathbf{PE} \text{ if } i=1,
\end{aligned}
\end{equation}
where $\mathbf{PE}$ represents a learnable positional encoding incorporated exclusively at the first stage to retain spatial inductive biases. This progressive compression ensures that features of extremely small targets are strictly preserved in the early, high-resolution stages. However, maintaining high-resolution feature maps intrinsically yields ultra-long 2D token sequences (e.g., ${\sim}100,000$ tokens), a computational bottleneck that is subsequently resolved by our dynamic sparse mixer.

\subsubsection{Dynamic Sparse Mixer} \label{sec:sparse-mixer}

To mitigate the quadratic or scaling constraints of modeling ultra-long sequences without uniformly dropping tokens, the Sparse Mixer block introduces a routing-based incremental modeling strategy (Fig. \ref{fig:vis_model_overview} b)). At the $i$-th stage, the feature map $\mathbf{F}_i$ is first flattened into a 1D sequence $\mathbf{s}_i \in \mathbb{R}^{L_i \times d_i}$, where the sequence length is $L_i = H_i \times W_i$.

The sequence is then iteratively refined by $N_i$ Adaptive Token Routing and Incremental Modeling (ATRIM) units, followed by an unflattening operation to restore the 2D spatial topology:
\begin{align}
\mathbf{s}_i &= \Phi_{\mathrm{flatten}}(\mathbf{F}_i), \\
\mathbf{s}_i &= \Phi_{\mathrm{ATRIM}}^{k}(\mathbf{s}_i, r_i), \quad k = 1, \dots, N_i, \\
\mathbf{F}_i &= \Phi_{\mathrm{un\text{-}flatten}}(\mathbf{s}_i).
\end{align}
Here, $\Phi_{\mathrm{ATRIM}}^{k}$ denotes the $k$-th unit within the $i$-th stage, and $r_i \in [0, 1)$ is a layer-specific token reduction ratio. The parameter $r_i$ dynamically adapts to the spatial redundancy level at each processing stage, effectively filtering background patches (e.g., homogeneous oceans or bare land) from the heavy computational stream.




\subsubsection{Adaptive Token Routing \& Incremental Modeling (ATRIM)} \label{sec:ATRIM}

Directly discarding visual tokens, a common practice in natural image perception, introduces severe inductive biases for remote sensing tasks by permanently fracturing the continuous spatial structure of roads, buildings, and land covers. To resolve this tension, ATRIM unit processes the input sequence $\mathbf{s}$ via three sequential operations: i) semantic token routing, ii) intensive dual-path SSM scanning exclusively on salient tokens, and iii) context-preserving incremental connection (Fig. \ref{fig:vis_model_overview} c)). The unit is formalized as:
\begin{equation}
\begin{aligned}
\mathbf{s} &= \Phi_{\mathrm{ATRIM}}(\mathbf{s}, r) \\
&= \Phi_{\mathrm{increment\text{-}conn}} \circ \Phi_{\mathrm{dual\text{-}scan}} \circ \Phi_{\mathrm{token\text{-}route}}(\mathbf{s}, r).
\end{aligned}
\end{equation}

\noindent \textbf{i) Semantic Token Routing:}
To guarantee both localized focus and macroscopic awareness, the router partitions the input sequence $\mathbf{s} \in \mathbb{R}^{L \times d}$ into two distinct token subsets: sparse regional semantics ($\mathbf{x}_r$) and dense global semantics ($\mathbf{x}_g$). The global semantics $\mathbf{x}_g \in \mathbb{R}^{\sqrt{L} \times d}$ are derived via adaptive 1D pooling, which compresses the sequence length from $L$ to $\sqrt{L}$ to ensure that the broader context is preserved.

For the regional semantics $\mathbf{x}_r$, a linear projection layer $\Phi_{\mathrm{mlp}}$ maps the input dimension $d$ to $1$ to compute the importance logits for every token. To encourage exploration during early training phases and prevent premature convergence to sub-optimal local minima, Gumbel-distributed noise is injected into these logits. The routing process, which identifies the top-$k$ most salient tokens according to a reduction ratio $r$, is formulated as follows:
\begin{align}
\mathbf{p} &= \Phi_{\mathrm{mlp}}(\mathbf{s}), \\
\mathbf{w}' &= \sigma\left(\mathbf{p} + \epsilon\right), \quad \epsilon \sim \mathrm{Gumbel}\left(0, \tau_e\right), \\
\Omega &= \Phi_{\mathrm{top\text{-}}k}(\mathbf{w}', r), \\
\mathbf{x}_r &= \mathbf{s}_{\Omega},
\end{align}
where $\mathbf{p} \in \mathbb{R}^{L \times 1}$ represents the unnormalized importance logits and $\sigma$ denotes the softmax function. The epoch-dependent noise temperature is defined as $\tau_e=\nu\left(1-e/E_{\mathrm{max}}\right)$, which progressively decays over the current training epoch $e$ relative to the maximum epochs $E_{\mathrm{max}}$, with $\nu=0.1$ serving as a scaling factor. The index set $\Omega$ contains the spatial indices of the retained tokens, and $\mathbf{s}_{\Omega}$ denotes the extraction of these tokens from the original sequence.

\begin{figure}[!t]
\centering
\includegraphics[width=1\linewidth]{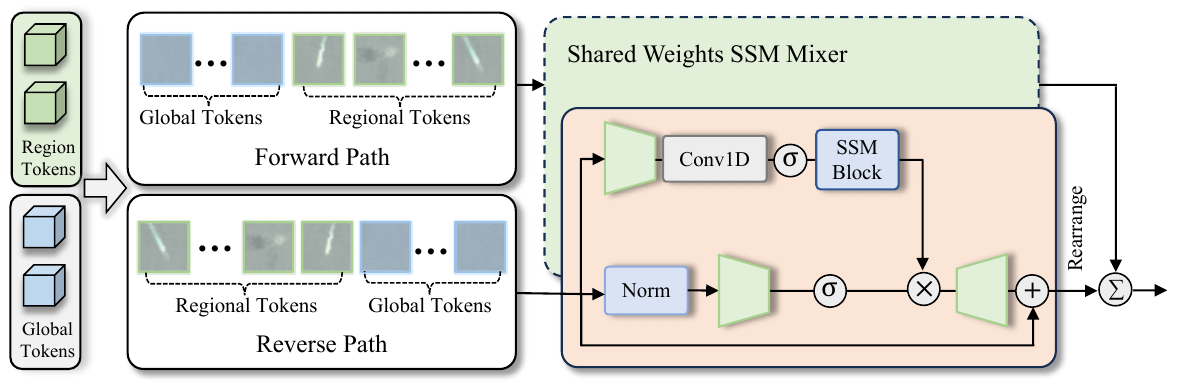}
\caption{
Architecture of the streamlined dual-path SSM scanning, applied exclusively to the routed salient regional tokens and the pooled global context.
} \label{fig:vis_model_dual_scanning}
\vspace{-0.3cm}
\end{figure}

\noindent \textbf{ii) Dual-Path SSM Scanning:}
Recent Vision Mamba models typically mandate 4-directional or 8-directional continuous scans across the entire spatial grid, resulting in high redundancy \cite{zhu2024vision,liu2024vmamba}. Conversely, our approach concatenates the pooled global tokens and the highly salient regional tokens, $[\mathbf{x}_g, \mathbf{x}_r]$, processing them through a streamlined dual-path (forward and backward) Mamba block \cite{chen2024rsmamba, gu2024mamba} (see Fig. \ref{fig:vis_model_dual_scanning}). Because the top-$k$ selection mechanism provides theoretically arbitrary token sequence orderings, non-causal modeling in Mamba is well-supported:
\begin{equation}
[\mathbf{x}'_g, \mathbf{x}'_r] = \Phi_{\mathrm{dual\text{-}scan}}([\mathbf{x}_g, \mathbf{x}_r]).
\end{equation}
This design restricts the $\mathcal{O}(N)$ state-space computation strictly to the most informative spatial elements, maximizing resource allocation for critical foreground targets.


\noindent \textbf{iii) Context-Preserving Incremental Connection:}
Crucially, unselected background tokens are not discarded. Instead, the refined sparse features ($\mathbf{x}'_r$) are conceptualized as incremental semantic updates. These updates are projected back to their original spatial positions and fused with the initial sequence using deterministic routing probabilities:
\begin{align}
\mathbf{w} &= \sigma(\mathbf{p}), \quad (\text{\footnotesize noise-free importance weights}), \label{eq:CPIC1} \\
\mathbf{s}' &= \mathbf{w} \odot \mathbf{s}, \quad (\text{\footnotesize importance-weighted original tokens}), \label{eq:CPIC2} \\
\mathbf{s}'_{\Omega} &= \mathbf{w}_{\Omega} \odot \mathbf{x}'_r, \quad (\text{\footnotesize in-place update at selected indices}), \label{eq:CPIC3} \\
\mathbf{s}_{\mathrm{out}} &= \mathbf{s} + \mathbf{s}', \quad (\text{\footnotesize residual integration}). \label{eq:CPIC4}
\end{align}
Eq. \ref{eq:CPIC1} defines $\mathbf{w}$ as the pure semantic relevance score (omitting Gumbel noise during fusion). In Eq. \ref{eq:CPIC2}, $\odot$ denotes element-wise multiplication broadcasted along the channel dimension. In Eq. \ref{eq:CPIC3}, the updated sparse tokens replace their corresponding spatial slots within the weighted sequence $\mathbf{s}'$. Finally, the residual connection in Eq. \ref{eq:CPIC4} reintegrates the computationally bypassed background tokens. This parameter-free reconstruction strictly preserves the continuous spatial topology required for dense pixel-level interpretations while saving massive computational overhead.


\subsection{Region-Level Meta-Embedding MIL Pre-training}

\subsubsection{Motivation \& Formulation}
A fundamental prerequisite for the success of the dynamic routing module is its capability to recognize and separate sparse foreground instances from massive, homogeneous backgrounds. Conventional contrastive learning provides only global image-level alignment, which easily overlooks tiny targets. Conversely, MAE pre-training forces the network to memorize background pixel textures, failing to instill the high-level semantic discriminability necessary for accurate routing.

\begin{figure}[!t]
\centering
\includegraphics[width=\linewidth]{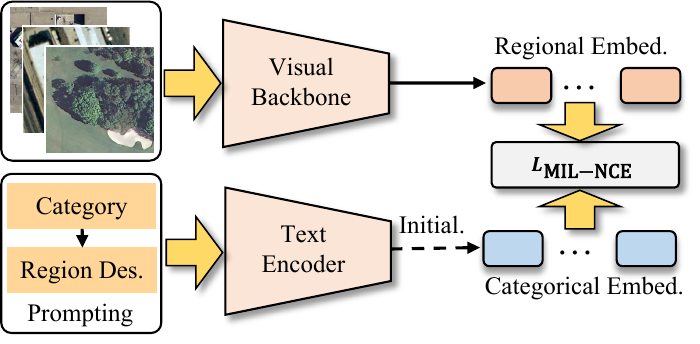}
\caption{
Overview of the Region-Level Meta-Embedding MIL Pre-training paradigm. By pulling regional visual embeddings toward their corresponding semantic meta-embeddings and repelling them from negatives in a shared latent space, the network is explicitly trained to decouple sparse foreground targets from redundant backgrounds.
} \label{fig:vis_training_mil}
\vspace{-0.3cm}
\end{figure}

To address this misalignment, we propose a Region-Level Meta-Embedding MIL paradigm. Utilizing the million-scale fMoW dataset \cite{christie2018functional} with its weak, noisy region-level annotations, we treat each massive image $\mathcal{I}_i$ as a ``bag'' of $n_i$ instances $y_i = \{\mathbf{b}_{i,j}, \mathbf{c}_{i,j}\}_{j=1}^{n_i}$, where $\mathbf{b}$ denotes bounding box coordinates and $\mathbf{c}$ denotes categorical semantic labels. By contrasting region-specific visual representations against categorical meta-embeddings in the latent space (Fig. \ref{fig:vis_training_mil}), the model explicitly learns to decouple heterogeneous foregrounds from the background. This establishes robust semantic priors that guide the dynamic router during fine-tuning.

The Multi-Instance Learning Noise Contrastive Estimation (MIL-NCE) loss, $\mathcal{L}_{\mathrm{MIL-NCE}}$, is formulated as:
{\small
\begin{equation}
\mathcal{L}_{\mathrm{MIL-NCE}} = -\log \frac{\sum_{(\mathbf{v}, \mathbf{t}) \in \mathcal{P}} f(\mathbf{v}, \mathbf{t})}{\sum_{(\mathbf{v}, \mathbf{t}) \in \mathcal{P}} f(\mathbf{v}, \mathbf{t}) + \sum_{(\mathbf{v}', \mathbf{t}') \in \mathcal{N}} f(\mathbf{v}', \mathbf{t}')},
\end{equation}
}%
where $f(\mathbf{x}, \mathbf{y}) = \exp \big( \langle \mathbf{x}, \mathbf{y} \rangle / \tau \big)$. Here, $\langle \cdot, \cdot \rangle$ denotes cosine similarity, $\mathcal{P}$ represents a set of positive matches between regional visual embeddings ($\mathbf{v}$) and their corresponding categorical meta-embeddings ($\mathbf{t}$), and $\mathcal{N}$ comprises negative pairs sampled from non-associated categories within the mini-batch. The parameter $\tau$ is a learnable temperature scalar.

\subsubsection{Regional Visual Representation}

Given an image $\mathcal{I}$, multi-level features $\{\mathbf{F}_i\}$ are extracted by the dynamic backbone. To ensure that the pre-training process remains agnostic to specific scales and robust to the inherently weak nature of fMoW annotations, we employ a Generic RoI Extractor (GRoIE) \cite{rossi2021novel} to derive scale-invariant visual representations for each bounding box $\mathbf{b}$. Instead of heuristically assigning an RoI to a single feature level, GRoIE aggregates features across all hierarchical scales:
\begin{equation}
\mathbf{v} = \Phi_{\mathrm{pool}} \left( \Phi_{\mathrm{g\text{-}roi}} \left( \{\mathbf{F}_i\}_{i=1}^5, \mathbf{b} \right) \right),
\end{equation}
where the output $\mathbf{v} \in \mathbb{R}^{1 \times d}$ is the localized visual embedding obtained after applying global spatial average pooling to the extracted multiscale region.

\subsubsection{Categorical Meta-Embedding via Language Priors}


Rather than relying on a traditional parameterized classification head, which rigidly restricts the model to fixed classes, we project the semantic labels into a continuous feature space as learnable points, termed \textit{meta-embeddings} ($\mathbf{t}$). To supply strong initial semantic priors and accelerate pre-training convergence, these meta-embeddings are initialized using text embeddings extracted from a pre-trained OpenCLIP text encoder \cite{radford2021learning,cherti2023reproducible}. We utilize zero-shot prompt templates from OpenCLIP’s standard protocol\footnote{https://github.com/mlfoundations/open\_clip} (e.g., ``An aerial image of a [CLASS]"). This strategy naturally aligns visual features with rich language semantics, promoting cross-task generalization and allowing the framework to scale beyond fixed categories to open-vocabulary region-text retrieval in future applications.

\subsection{Adaptation to Downstream Tasks}

Thanks to its standard hierarchical FPN architecture and the robust discriminative priors acquired during MIL pre-training, DynamicVis seamlessly adapts to a broad spectrum of downstream applications utilizing standard, unmodified task-specific decoders.

\noindent \textbf{Region-Level Interpretation} (e.g., scene classification, region classification, image retrieval) utilizes global average pooling on the highest-level semantic feature map. This produces compact embedding vectors suitable for linear classification heads or zero-shot similarity matching.

\noindent \textbf{Instance-Level Interpretation} (e.g., object detection, instance segmentation) couples the backbone with standard two-stage frameworks such as Faster R-CNN \cite{ren2016faster} and Mask R-CNN \cite{he2017mask}. The dynamic routing efficiently filters background noise early in the network, significantly enhancing the recall rates of small and densely clustered targets.

\noindent \textbf{Pixel-Level Interpretation} (e.g., semantic segmentation, change detection) relies heavily on the uninterrupted spatial structure ensured by our incremental residual connections. This intact topology enables dense prediction heads, such as UperNet \cite{xiao2018unified} or bi-temporal difference MLPs, to achieve precise edge delineation and structural connectivity.

\section{Experiments and Analysis}
\label{sec:experiments}

To comprehensively validate the efficacy of the proposed DynamicVis framework, we conduct extensive experiments designed to address three core research questions:
\textbf{i) Semantic Decoupling:} Can the Region-Level Meta-Embedding MIL pre-training paradigm effectively decouple sparse foreground instances from massive, redundant backgrounds?
\textbf{ii) Sparse-Target Superiority:} Does the dynamic token routing mechanism successfully translate into superior performance on tasks requiring sparse-target perception (e.g., small object detection) without compromising the structural integrity required for continuous dense prediction capabilities?
\textbf{iii) Computational Scalability:} How does DynamicVis scale in terms of memory consumption and inference latency when processing ultra-high-resolution remote sensing imagery compared to conventional ViTs and dense SSMs?

\subsection{Pre-training Configuration and Implementation Details}

\subsubsection{Pre-training Dataset: fMoW}

Unlike conventional MAE pre-training paradigms that indiscriminately reconstruct pixel-level background noise, our dynamic routing module necessitates explicit, high-level semantic guidance to accurately distinguish foreground from background. To fulfill this prerequisite, we utilize the Functional Map of the World (fMoW) dataset \cite{christie2018functional}.

\begin{figure}[!t]
\centering
\includegraphics[width=\linewidth]{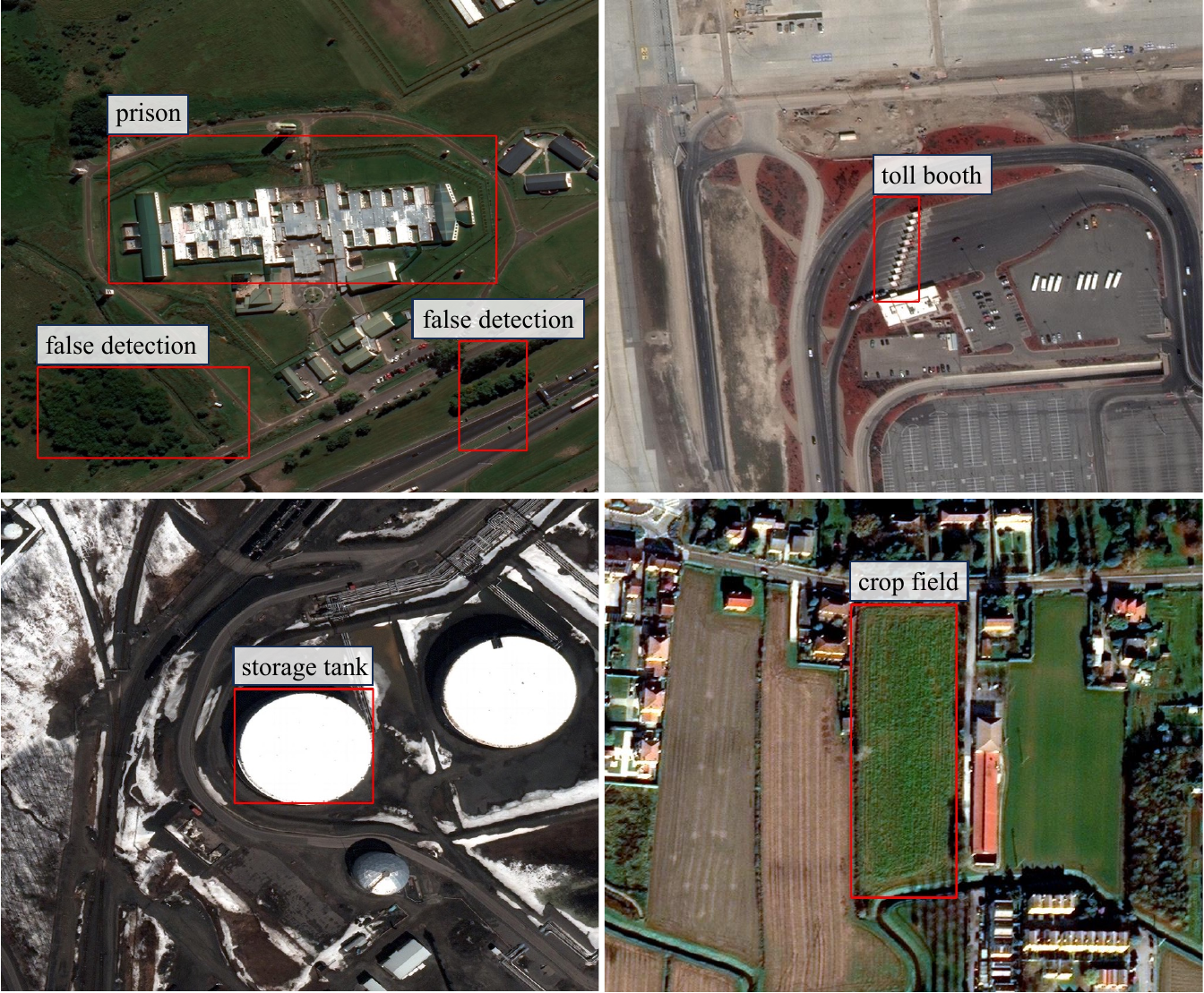}
\caption{
Illustrative samples from the fMoW dataset \cite{christie2018functional}. The extreme variations in scale, geographical location, and temporal conditions provide a robust foundation for the Meta-Embedding MIL pre-training paradigm, compelling the model to learn invariant, high-level semantic representations of targets.
} \label{fig:fig_dataset_samples}
\vspace{-0.3cm}
\end{figure}

Acquired via the DigitalGlobe satellite constellation, fMoW serves as a critical bridge between global image classification and precise object detection. It provides weak, loosely defined bounding box annotations for 62 distinct semantic categories, alongside an auxiliary ``false detection'' class for uncategorized instances. Spanning over 200 countries, the dataset encompasses diverse geographical and temporal distributions, as illustrated in Fig. \ref{fig:fig_dataset_samples}. For our pre-training regimen, we adopt the fMoW-rgb version, consolidating the training, validation, and sequence splits into a massive corpus of 1,027,691 training images. A hold-out set of 20,000 test images is utilized exclusively to monitor pre-training convergence and to evaluate zero-shot region classification. By treating these massive, heterogeneous scenes as ``bags'' of instances, our MIL paradigm compels the network to explicitly map foreground targets into a structured, separable latent semantic space, fundamentally overcoming the semantic ambiguity of pixel-reconstruction approaches.

\subsubsection{Architectural Specifications}

\begin{table*}[!htb]
\centering
\caption{Architectural specifications of the \textit{DynamicVis} configurations. The progressive downsampling strictly preserves fine-grained local details, while the stage-wise reduction ratio ($r_i$) progressively retains more tokens in deeper layers to maintain the macroscopic spatial topology required for dense downstream tasks.}
\label{tab:model_architecture}
\setlength{\tabcolsep}{4pt}
\begin{tabular}{l| ccccc}
\toprule
Version & Layers ($N_i$) & Dimensions ($d_i$) & Reduction Ratio ($r_i$) & Patch Size ($k_i$) & Stride ($s_i$) \\
\midrule
DynamicVis-Base & $[2, 4, 16, 4]$ & $[96, 192, 384, 768]$ & $[7/8, 3/4, 1/2, 0]$ & $[7, 3, 3, 3]$ & $[4, 2, 2, 2]$ \\
DynamicVis-Large & $[2, 4, 32, 4]$ & $[128, 256, 512, 1024]$ & $[7/8, 3/4, 1/2, 0]$ & $[7, 3, 3, 3]$  & $[4, 2, 2, 2]$ \\
\bottomrule
\end{tabular}
\end{table*}

To evaluate the architectural scalability of our approach, we instantiate DynamicVis in two distinct configurations: \textbf{Base} and \textbf{Large}. The structural hyperparameters across the four hierarchical processing stages are systematically detailed in Table \ref{tab:model_architecture}.

Crucially, rather than adopting the aggressive $16 \times 16$ patch embedding typical of ViTs, which irrevocably destroys fine-grained spatial inductive biases, DynamicVis employs a progressive, small-stride downsampling strategy (patch sizes of $[7, 3, 3, 3]$ and strides of $[4, 2, 2, 2]$), thereby preserving the granular geometric configurations vital for localizing tiny objects.

Simultaneously, the dynamic sparse mixer applies stage-wise token reduction ratios of $[7/8, 3/4, 1/2, 0]$. This progressive sparsification strategy drops highly redundant background tokens (e.g., expansive oceans or bare land) early in the network to maximize computational efficiency, while retaining all tokens in the deepest layer to preserve the macroscopic semantic structures required for complex interpretation. Feature representations are computed via a bidirectional Mamba scanning module that aggregates outputs from forward and reverse scans, coupled with randomly initialized learnable positional embeddings. A standard FPN seamlessly projects these outputs into five multi-scale feature maps with a uniform dimension of 256 channels, ensuring direct compatibility with standard, unmodified dense prediction heads.

\subsubsection{Pre-training Protocols and Implementation}

To alleviate I/O bottlenecks in massive-scale pre-training, we formatted the fMoW dataset into 128 WebDataset-compliant\footnote{https://github.com/webdataset/webdataset} tar archives, enabling streamlined sequential data access.

All pre-training images were uniformly resampled to $512 \times 512$ pixels, with standard data augmentations including random horizontal flipping, resizing, and cropping. Optimization was performed using the AdamW optimizer with an initial learning rate of $4 \times 10^{-4}$ and a cosine annealing schedule over 200 epochs (requiring approximately 3,000 GPU hours on 8 NVIDIA A100 GPUs). Batch sizes were set to 512 and 256 for the Base and Large variants, respectively. The framework was implemented in PyTorch utilizing the OpenMMLab\footnote{https://openmmlab.com/codebase} ecosystem. To maximize computational throughput, Automatic Mixed Precision (AMP) in BF16 format was consistently applied, and distributed feature aggregation across nodes was utilized prior to loss computation to ensure the stability and convergence of the MIL objective.

A dual-objective loss function was employed, integrating the MIL-NCE objective with standard classification cross-entropy. To encourage initial exploration of salient features without relying on auxiliary balancing losses, Gumbel-distributed noise was injected into the routing affinity scores. The intensity of this noise was progressively annealed throughout training, ensuring a seamless transition to deterministic, optimal token routing during downstream inference.

\subsubsection{Evaluation Variants}

To rigorously isolate the performance contributions of our proposed architectural and training innovations, we define three model variants evaluated consistently across downstream tasks:
i) \textbf{DynamicVis$^\dag$}: The baseline dense SSM backbone. It processes all tokens uniformly (no adaptive routing) and is trained via plain cross-entropy without the MIL pre-training.
ii) \textbf{DynamicVis$^\ddag$}: The structurally complete architecture featuring adaptive token routing and incremental residual connections, but without pre-training of the MIL paradigm.
iii) \textbf{DynamicVis}: The full, finalized framework integrating both the dynamic region-aware token routing and the robust semantic priors established by the MIL pre-training.

\subsection{Evaluation on Sparse and Instance-Level Perception} \label{subsec:sparse_instance_evaluation}

We first evaluate DynamicVis on downstream tasks fundamentally characterized by target sparsity and discrete instance boundaries. These tasks serve as the primary testbed for validating our core hypothesis: adaptively routing computational resources exclusively to highly salient foreground regions significantly outperforms the dense, uniform processing paradigms typical of ViTs.

\subsubsection{Tiny Ship Detection}

\begin{table*}[!tbp]
\centering
\begin{minipage}[t]{0.28\linewidth}
\centering
\caption{
Quantitative comparison of tiny ship detection performance on the LEVIR-Ship dataset.
}\label{tab:levirship_sota}
\resizebox{1\linewidth}{!}{
\begin{tabular}{l c c c}
\toprule
Method & $\text{AP50}$ & $\text{AP50}_{\text{s}}$ & $\text{AP50}_{\text{m}}$ \\
\midrule
Faster R-CNN \cite{ren2016faster} & 75.2  & 74.5 & 86.5 \\
SSD \cite{liu2016ssd} & 56.9  & 55.8 & 63.4 \\
YOLOv3 \cite{redmon2018yolov3} & 70.7 & 71.4 & 56.8 \\
YOLOv5s \cite{yolov5} & 75.6 & - & - \\
TridentNet \cite{li2019scale}  & 75.1 & - & - \\
FCOS \cite{tian2020fcos}  & 80.4 & 79.8 & 82.2 \\
RetinaNet \cite{lin2017focal} & 75.3 & 72.3 & 75.0 \\
YOLOX \cite{ge2021yolox} & 81.9 & \cellcolor{gray!10}\textbf{81.2} & 89.1 \\
YOLOF \cite{chen2021you} & 68.1 & 66.6 & 85.9 \\
SparseRCNN \cite{sun2021sparse} & 71.9 & 71.3 & 73.4 \\
RTMDet \cite{lyu2022rtmdet} & 75.0 & 73.6 & 89.1 \\
EfficientDet \cite{tan2020efficientdet}  & 75.0 & 73.3 & 90.2 \\
CenterNet \cite{zhou2019objects} & 76.2 & 74.4 & 80.2 \\
ViTDet \cite{li2022exploring} & 75.3 & 73.6 & 90.9 \\
DeformableDETR \cite{zhu2020deformable} & 78.6 & 78.5 & 75.6 \\
DINO \cite{zhang2022dino} & 79.6 & 78.8 & 91.2 \\
CoDETR \cite{zong2023detrs} & 82.1  & 80.5 & 95.0 \\
DRENet \cite{chen2022degraded} & 82.4 & - & - \\
HSFNet \cite{li2018hsf} & 73.6 & - & - \\
ImYOLOv3 \cite{chen2021improved} & 72.6 & - & - \\
IM-YOLOv5s \cite{jian2023optical} & 76.9 & - & - \\
DS-YOLOv5s  \cite{huang2024deep}& 75.4 & - & - \\
DSFPAPNet \cite{jiang2024dsfpap}  & 82.6 & - & - \\
ORFENet \cite{liu2024tiny} & 83.3 & - & - \\
HRNet \cite{wang2020deep} & \cellcolor{gray!30}\textbf{83.7} & - & - \\
\midrule
DynamicVis-B$^\dag$ & 78.2 & 78.4 & 87.7 \\
DynamicVis-B$^\ddag$ & 80.3 & 79.0 & 95.3 \\
DynamicVis-B & \cellcolor{gray!10}\textbf{83.5} & \cellcolor{gray!30}\textbf{82.1} & \cellcolor{gray!70}\textbf{97.3} \\
DynamicVis-L$^\dag$ & 80.1 & 79.9 & 82.9 \\
DynamicVis-L$^\ddag$ & 82.3 & 80.8 & \cellcolor{gray!10}\textbf{95.5} \\
DynamicVis-L & \cellcolor{gray!70}\textbf{84.1} & \cellcolor{gray!70}\textbf{82.8} & \cellcolor{gray!30}\textbf{96.8} \\
\bottomrule
\end{tabular}
}
\end{minipage}
\hfill
\begin{minipage}[t]{0.347\linewidth}
\centering
\caption{
Comparison of road extraction performance on the Massachusetts dataset.
}\label{tab:massachusetts_sota}
\resizebox{1\linewidth}{!}{
\begin{tabular}{l c c c c}
\toprule
Method & P & R & F1 & IoU \\
\midrule
SegNet \cite{badrinarayanan2017segnet} & 78.89 & 67.73 & 72.25 & 57.02 \\
U-Net \cite{ronneberger2015u}  & 77.53 & 77.82 & 77.67 & 63.50 \\
ResUNet \cite{diakogiannis2020resunet}  & 80.76 & 71.49 & 75.69 & 61.21 \\
D-LinkNet \cite{zhou2018d}  & 78.34 & \cellcolor{gray!10}\textbf{77.91} & 78.12 & 64.10 \\
HRNetv2 \cite{wang2020deep} & 79.01 & \cellcolor{gray!30}\textbf{78.20} & 78.60 & 64.75 \\
DeeplabV3 \cite{chen2017rethinking}  & 80.81 & 75.17 & 77.89 & 63.79 \\
DANet \cite{fu2019dual} & 81.69 & 75.39 & 78.41 & 64.49 \\
DeeplabV3+ \cite{chen2018encoder} & 83.07 & 75.62 & 79.17 & 65.53 \\
Mask2Former \cite{cheng2022masked} & 80.17 & 73.87 & 76.63 & 62.11 \\
HRNet \cite{wang2020deep} & 83.57 & 75.78 & 79.49 & 65.96 \\
PoolFormer \cite{yu2022metaformer} & 83.16 & 74.99 & 78.87 & 65.11 \\
PSPNet \cite{zhao2017pyramid} & 82.77 & 73.15 & 77.66 & 63.48 \\
PSANet \cite{zhao2018psanet} & 80.63 & 76.19 & 78.35 & 64.41 \\
UperNet \cite{xiao2018unified} & 83.03 & 75.90 & 79.30 & 65.70 \\
Segformer \cite{xie2021segformer} & 83.55 & 74.78 & 78.92 & 65.18 \\
SIINet \cite{tao2019spatial}  & \cellcolor{gray!70}\textbf{85.36} & 74.13 & 79.35 & 65.77 \\
BDTNet \cite{luo2022bdtnet} & 82.99 & 76.37 & 79.54 & 66.03 \\
RoadFormer \cite{jiang2022roadformer}  & 80.54 & \cellcolor{gray!70}\textbf{78.90} & 79.71 & 66.27 \\
GA-Net \cite{chen2022ga} & 84.10 & 76.89 & \cellcolor{gray!30}\textbf{80.33} & \cellcolor{gray!30}\textbf{67.13} \\
\midrule
DynamicVis-B$^\dag$  & 83.67 & 73.55 & 78.34 & 64.32 \\
DynamicVis-B$^\ddag$ &  83.79 & 71.94 & 78.09 & 63.16 \\
DynamicVis-B & 82.31 & 76.44 & 79.26 & 65.66 \\
DynamicVis-B$^{2\times}$ &  83.27 & 76.94 & \cellcolor{gray!10}\textbf{79.87} &  \cellcolor{gray!10}\textbf{66.56} \\
DynamicVis-L$^\dag$  & \cellcolor{gray!10}\textbf{84.12} & 73.72 & 78.57 & 64.70 \\
DynamicVis-L$^\ddag$  & 82.82 & 73.87 & 78.23 & 64.06 \\
DynamicVis-L & 83.18 & 76.52 & 79.72 &  66.26 \\
DynamicVis-L$^{2\times}$ & \cellcolor{gray!30}\textbf{85.05} & 76.27 & \cellcolor{gray!70}\textbf{80.35} & \cellcolor{gray!70}\textbf{67.20} \\
\bottomrule
\end{tabular}
}
\end{minipage}
\hfill
\begin{minipage}[t]{0.335\linewidth}
\centering
\caption{Comparison of building extraction performance on the WHU dataset.
}\label{tab:whu_building_sota}
\resizebox{1\linewidth}{!}{
\begin{tabular}{l c c c c}
\toprule
Method & P & R & F1 & IoU \\
\midrule
FCN \cite{long2015fully} & 92.29 & 92.84 &  92.56 &  86.16 \\
SegNet \cite{badrinarayanan2017segnet}  & 93.42&  91.71 & 92.56&  86.15 \\
U-Net \cite{ronneberger2015u} &  94.50 & 90.88 & 92.65 & 86.31 \\
MA-FCN \cite{chang2024multi} &  94.75 & 94.92 & 94.83 & 90.18 \\
DeepLabv3 \cite{chen2017rethinking} & 95.03 & 93.12 & 94.06 & 88.80 \\
Deeplabv3+ \cite{chen2018encoder}  & 94.31 & 94.53 & 94.42 & 89.43 \\
DANet \cite{fu2019dual} & 95.13 & 94.12 & 94.62 & 89.80 \\
Mask2Former \cite{cheng2022masked} & 92.26 & 92.22 & 92.24 & 85.60 \\
HRNet \cite{wang2020deep} & 94.78 & 93.64 & 94.20 & 89.05 \\
PoolFormer \cite{yu2022metaformer} & 95.10 & 94.24 & 94.67 & 89.88 \\
PSPNet \cite{zhao2017pyramid} & 94.46 & 94.38 & 94.42 & 89.43 \\
UperNet \cite{xiao2018unified} & 95.60 & 94.21 & 94.90 & 90.30 \\
PSANet \cite{zhao2018psanet} & 94.15 & 93.96 & 94.06 & 88.78 \\
ResUNet \cite{diakogiannis2020resunet} & 94.49 & 94.71 & 94.60 & 89.75 \\
MAP-Net  \cite{zhu2020map} &  93.99&  94.82&  94.40&  89.40 \\
Segformer \cite{xie2021segformer}  &  94.72 & 94.42 & 94.57 & 89.70 \\
TransUNet \cite{zhang2021transformers} &  94.05 & 93.07 & 93.56 & 87.89 \\
CMTFNet \cite{wu2023cmtfnet}  & 90.12 & \cellcolor{gray!30}\textbf{95.21} & 92.59 & 86.21 \\
RSM-SS \cite{zhao2024rs}  & 95.25 & \cellcolor{gray!10}\textbf{95.12} & 95.18 & 90.81 \\
STT \cite{chen2021building} & - & - & 94.97 & 90.48 \\
\midrule
DynamicVis-B$^\dag$  &95.59 & 94.06 & 94.83& 90.16 \\
DynamicVis-B$^\ddag$  & 94.91 & 93.83 & 94.37 & 89.33 \\
DynamicVis-B &  95.43 & 95.11 & 95.27 & 90.98 \\
DynamicVis-B$^{2\times}$ & \cellcolor{gray!10}\textbf{95.82} & 95.01 & \cellcolor{gray!30}\textbf{95.41} & \cellcolor{gray!30}\textbf{91.23} \\
DynamicVis-L$^\dag$  &  95.36 & 94.72 & 95.04 & 90.55 \\
DynamicVis-L$^\ddag$  &  94.45 & 94.17 & 94.31 & 89.24 \\
DynamicVis-L &  \cellcolor{gray!70}\textbf{96.39} & 94.27 & \cellcolor{gray!10}\textbf{95.32} & \cellcolor{gray!10}\textbf{91.06} \\
DynamicVis-L$^{2\times}$ & \cellcolor{gray!30}\textbf{95.87} & \cellcolor{gray!70}\textbf{95.28} & \cellcolor{gray!70}\textbf{95.58} & \cellcolor{gray!70}\textbf{91.54} \\
\bottomrule
\end{tabular}
}
\end{minipage}
\end{table*}

\textbf{Datasets \& Implementations:} Ship detection in remote sensing imagery presents a formidable challenge, as targets are exceptionally small and frequently obfuscated by environmental interference (e.g., clouds, waves, and reefs). We utilize the LEVIR-Ship dataset \cite{chen2022degraded}, comprising 3,896 high-resolution optical images ($512 \times 512$) acquired by GF-1 and GF-6 satellites at a 16m spatial resolution. Target sparsity is extreme: ship instances average merely $10 \times 10$ pixels, with approximately $90\%$ occupying less than $20 \times 20$ pixels. Incorporating DynamicVis as the backbone within an FCOS \cite{tian2020fcos} detection framework, the network is fine-tuned for 200 epochs using the AdamW optimizer (initial learning rate of $2 \times 10^{-4}$ and batch size of 64). Evaluation strictly follows COCO metrics, focusing on $\text{AP50}$, small object precision ($\text{AP50}_{\text{s}}$ for objects $<32 \times 32$), and medium object precision ($\text{AP50}_{\text{m}}$ for objects $32 \times 32 \sim 64 \times 64$).

\textbf{Results and Analysis:} As detailed in Table \ref{tab:levirship_sota}, DynamicVis-L establishes a new state-of-the-art, particularly dominating the critical $\text{AP50}_{\text{s}}$ metric for tiny targets. Three key insights emerge:
\textit{i) Mitigation of Feature Dilution:} Traditional ViTs uniformly allocate attention across the vast, target-free ocean, which inadvertently dilutes the feature representations of miniature ships. Conversely, our dynamic token routing mechanism (evidenced by the $+2.2$ point leap in $\text{AP50}$ from the dense baseline DynamicVis-L$^\dag$ to the routed DynamicVis-L$^\ddag$) explicitly prioritizes sparse foreground signals, preserving vital target fidelity.
\textit{ii) Efficacy of Anchor-Free Paradigms:} For sparsely distributed tiny objects, anchor-free detectors (like FCOS \cite{tian2020fcos}) exhibit superior performance by circumventing the optimization bottlenecks associated with matching thousands of redundant anchors to minute target regions.
\textit{iii) Global Context Utilization:} Transformer-based architectures (e.g., DETR variants \cite{carion2020end}) demonstrate surprisingly robust performance. Despite target diminutive size, their capacity to model global contextual relationships provides crucial discriminative priors for distinguishing ships from visually similar background clutter.

\begin{figure*}[!htbp]
\centering
\includegraphics[width=0.95\linewidth]{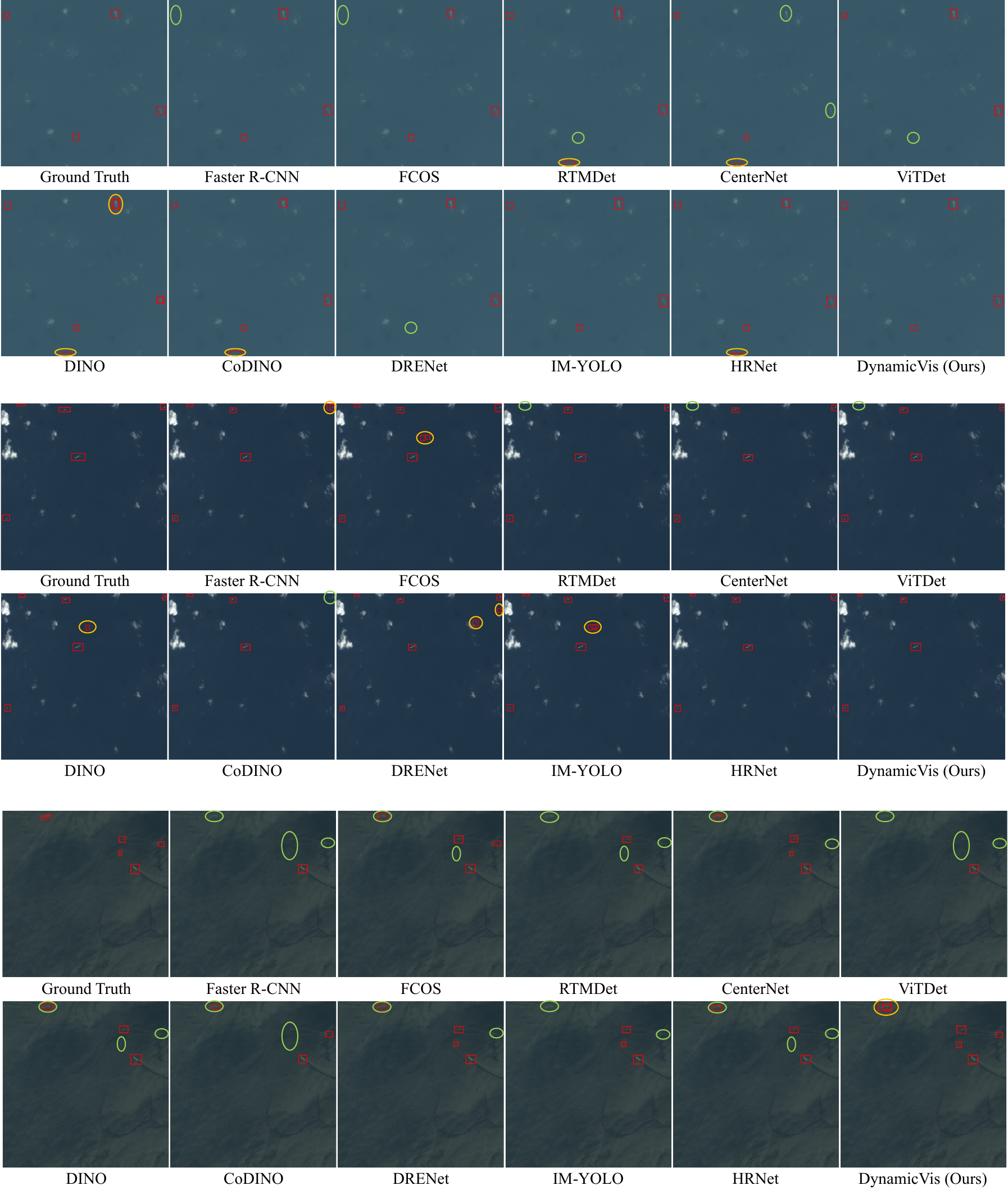}
\caption{
Qualitative detection performance comparison between DynamicVis-L and baseline architectures on the LEVIR-Ship test set. Annotations correspond to true positives (\textcolor[HTML]{D61215}{\textbf{red}}), false positives (\textcolor[HTML]{FAC000}{\textbf{yellow}}), and false negatives (\textcolor[HTML]{92D050}{\textbf{green}}).
} \label{fig:vis_levirship_comparision}
\end{figure*}

\begin{figure}[!tbp]
\centering
\includegraphics[width=\linewidth]{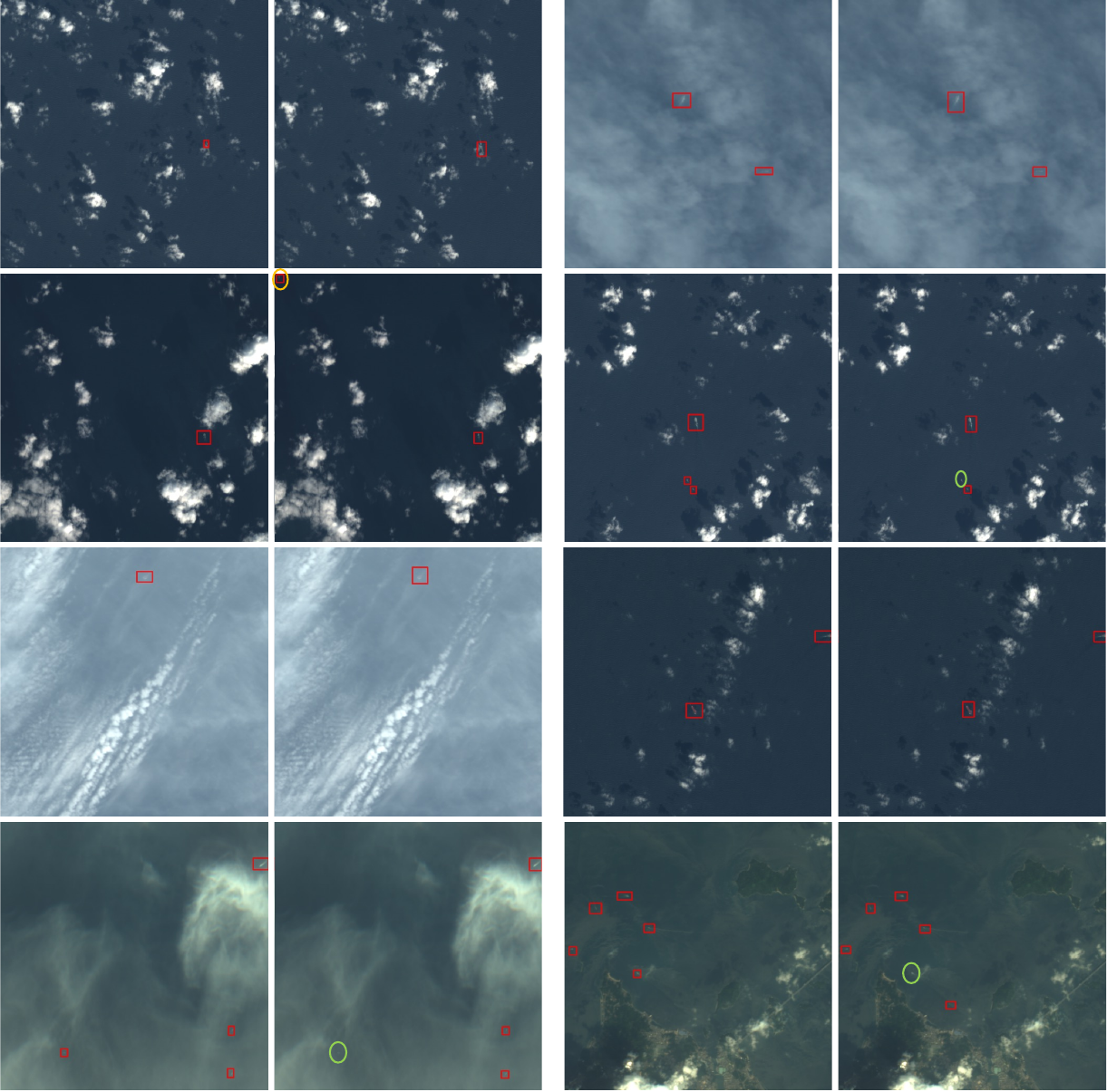}
\caption{
DynamicVis-L tiny ship detection examples under severe environmental interference on the LEVIR-Ship test set. Left: ground-truth annotations; Right: the corresponding model predictions.
} \label{fig:vis_levirship}
\end{figure}

\textbf{Visualizations:} As illustrated in Fig. \ref{fig:vis_levirship_comparision} and Fig. \ref{fig:vis_levirship}, DynamicVis exhibits exceptional robustness against severe environmental interference, such as cloud occlusion. By selectively bypassing background patches rather than uniformly merging them, the network drastically reduces both false negatives and spurious detections, validating our structure-preserving dynamic routing strategy.

\subsubsection{Instance Segmentation}

\textbf{Datasets \& Implementations:} Instance segmentation uniquely requires both precise spatial localization and pixel-perfect boundary delineation. We systematically evaluate on two complementary benchmarks: NWPU VHR-10 \cite{cheng2014multi} (comprising 10 highly diverse optical categories, such as bridges, vehicles, and harbors) and SSDD \cite{zhang2021sar} (focused on SAR-based ship detection, presenting severe modality-specific noise). We embed DynamicVis into a standard Mask R-CNN \cite{he2017mask} framework for 400 epochs. Model efficacy is rigorously quantified via COCO metrics for bounding box ($\text{AP}_{\text{box}}$) and mask predictions ($\text{AP}_{\text{mask}}$).

\textbf{Results and Analysis:} Comprehensive quantitative evaluations in Tables \ref{tab:nwpu_sota} and \ref{tab:ssdd_sota} reveal that DynamicVis significantly outperforms dense ViTs (e.g., Mask2Former \cite{cheng2022masked}) and matches or strictly exceeds highly specialized SAM-based foundation models like RSPrompter \cite{chen2024rsprompter}. Crucially, while RSPrompter relies on freezing a massive, pre-trained SAM encoder, which inherently limits domain adaptability, DynamicVis demonstrates exceptional flexibility and multi-modal robustness. This is most vividly illustrated on the SAR-based SSDD dataset, where DynamicVis-L achieves a dominant $71.0 \text{ AP}_{\text{mask}}$, outperforming RSPrompter-query ($67.3$) by $+3.7$ points.

\begin{table*}[!ht]
\centering
\begin{minipage}[t]{0.346\linewidth}
\centering
\caption{
Comparison of instance segmentation performance on the NWPU VHR-10 dataset.
}\label{tab:nwpu_sota}
\resizebox{1\linewidth}{!}{
\begin{tabular}{l c c c c c c}
\toprule
\multirow{2}{*}{Method} & \multicolumn{3}{c}{Detection} & \multicolumn{3}{c}{Segmentation} \\
\cmidrule(lr){2-4} \cmidrule(lr){5-7}
& $\text{AP}_{\text{box}}$ & $\text{AP}_{\text{box}}^{50}$ & $\text{AP}_{\text{box}}^{75}$
& $\text{AP}_{\text{mask}}$ & $\text{AP}_{\text{mask}}^{50}$ & $\text{AP}_{\text{mask}}^{75}$ \\
\midrule
Mask R-CNN \cite{he2017mask} & 62.3 & 88.3 & 75.2 & 59.7 & 89.2 & 65.6 \\
MS R-CNN \cite{huang2019mask} & 62.3 & 88.6 & 73.1 & 60.7 & 88.7 & 67.7 \\
HTC \cite{chen2019hybrid} & 63.9 & 88.9 & 75.4 & 60.9 & 88.6 & 64.4 \\
SOLO v2 \cite{wang2020solov2} & - & - & - &  50.9 & 77.5 & 54.1 \\
SCNet \cite{vu2021scnet} & 60.0 & 87.5 & 69.1 & 58.1 & 87.4 & 62.0 \\
CondInst \cite{tian2020conditional} & 62.3 & 87.8 & 73.3 & 59.0 & 88.5 & 62.8 \\
BoxInst \cite{tian2021boxinst} & 64.8 & 89.3 & 73.0 & 47.6 & 77.2 & 51.3 \\
Mask2Former \cite{cheng2022masked} & 57.4 & 75.5 & 63.7 & 58.8 & 83.1 & 63.5 \\
CATNet \cite{liu2024learning} & 63.2 & 89.0 & 73.8 & 60.4 & 89.6 & 65.5 \\
HQ-ISNet \cite{su2020hq} & 63.5 & 89.9 & 75.0 & 60.4 & 89.6 & 64.1 \\
\midrule
RSPrompter-anchor \cite{chen2024rsprompter} & \cellcolor{gray!70}\textbf{70.3} & \cellcolor{gray!70}\textbf{93.6} & \cellcolor{gray!70}\textbf{81.0} & 66.1 & \cellcolor{gray!70}\textbf{92.7} & 70.6 \\
RSPrompter-query \cite{chen2024rsprompter} & 68.4 & 90.3 & 74.0  & \cellcolor{gray!30}\textbf{67.5} & \cellcolor{gray!10}\textbf{91.7} & \cellcolor{gray!30}\textbf{74.8} \\
\midrule
DynamicVis-B$^\dag$ & 63.7 & 88.2 & 71.2 & 62.4 & 87.8 & 68.3 \\
DynamicVis-B$^\ddag$ & 65.0 & 89.9 & 71.9 & 63.2 & 88.8 & 67.8 \\
DynamicVis-B & \cellcolor{gray!10}\textbf{68.5} & \cellcolor{gray!10}\textbf{90.8} & \cellcolor{gray!10}\textbf{79.6} & \cellcolor{gray!10}\textbf{67.3} & 91.5 & \cellcolor{gray!10}\textbf{73.8} \\
DynamicVis-L$^\dag$ & 64.4 & 89.6 & 75.9 & 64.8 & 90.2 & 68.8 \\
DynamicVis-L$^\ddag$ & 64.9 & 89.8 & 75.5 & 65.1 & 90.8 & 69.2 \\
DynamicVis-L & \cellcolor{gray!30}\textbf{69.1} & \cellcolor{gray!30}\textbf{93.1} & \cellcolor{gray!30}\textbf{80.8} & \cellcolor{gray!70}\textbf{67.8} & \cellcolor{gray!30}\textbf{91.9} & \cellcolor{gray!70}\textbf{75.1} \\
\bottomrule
\end{tabular}
}
\end{minipage}
\hfill
\begin{minipage}[t]{0.346\linewidth}
\centering
\caption{
Comparison of instance segmentation performance on the SSDD dataset.
}\label{tab:ssdd_sota}
\resizebox{1\linewidth}{!}{
\begin{tabular}{l c c c c c c}
\toprule
\multirow{2}{*}{Method} & \multicolumn{3}{c}{Detection} & \multicolumn{3}{c}{Segmentation} \\
\cmidrule(lr){2-4} \cmidrule(lr){5-7}
& $\text{AP}_{\text{box}}$ & $\text{AP}_{\text{box}}^{50}$ & $\text{AP}_{\text{box}}^{75}$
& $\text{AP}_{\text{mask}}$ & $\text{AP}_{\text{mask}}^{50}$ & $\text{AP}_{\text{mask}}^{75}$ \\
\midrule
Mask R-CNN \cite{he2017mask} & 67.7 & 95.6 & 84.9 & 64.3 & 92.6 & 80.9 \\
MS R-CNN \cite{huang2019mask}  & 67.8 & 95.3 & \cellcolor{gray!10}\textbf{85.9} & 64.9 & 93.3 & 80.4 \\
HTC \cite{chen2019hybrid}  & 69.3 & 97.1 & 85.7 & 64.1 & 94.4 & 80.6 \\
SOLO v2 \cite{wang2020solov2} & - & - & - & 58.5 & 86.2 & 74.0 \\
SCNet \cite{vu2021scnet} & 66.9 & 92.5 &82.5 & 64.9 & 92.6 &80.1 \\
CondInst \cite{tian2020conditional} & 68.1 & 92.4 & 85.5 & 62.5 & 93.4 & 81.2 \\
BoxInst \cite{tian2021boxinst} & 62.8 & 96.2 & 74.7 & 45.2 & 92.3 & 35.3 \\
Mask2Former \cite{cheng2022masked}  & 62.7 & 90.7 & 75.6 & 64.4 & 93.0 & 82.4 \\
CATNet \cite{liu2024learning} & 67.5 & 96.8 & 80.4 & 63.9 & 93.7 & 80.1 \\
HQ-ISNet \cite{su2020hq} & 66.6 & 95.9 & 80.2 & 63.4 & 95.1 & 78.1 \\
\midrule
RSPrompter-anchor \cite{chen2024rsprompter} & \cellcolor{gray!10}\textbf{70.4} & \cellcolor{gray!30}\textbf{97.7} & \cellcolor{gray!30}\textbf{86.2} & 66.8 & 94.7 & 84.0 \\
RSPrompter-query \cite{chen2024rsprompter} & 66.0 &  95.6 & 78.7 & 67.3 & 95.6 & 84.3 \\
\midrule
DynamicVis-B$^\dag$ & 66.6 & \cellcolor{gray!70}\textbf{97.8} & 78.0 & 68.3 & \cellcolor{gray!70}\textbf{97.8} & 85.3 \\
DynamicVis-B$^\ddag$ & 67.0 & 97.6 & 81.3 & 68.3 & 97.5 & 86.2 \\
DynamicVis-B & \cellcolor{gray!30}\textbf{70.8} & \cellcolor{gray!30}\textbf{97.7} & \cellcolor{gray!70}\textbf{87.1} & \cellcolor{gray!30}\textbf{70.8} & \cellcolor{gray!30}\textbf{97.7} & \cellcolor{gray!70}\textbf{89.4} \\
DynamicVis-L$^\dag$ & 67.7 & \cellcolor{gray!30}\textbf{97.7} & 82.0 & 69.6 & \cellcolor{gray!30}\textbf{97.7} & \cellcolor{gray!10}\textbf{88.3} \\
DynamicVis-L$^\ddag$ & 68.9 & \cellcolor{gray!30}\textbf{97.7} & 84.2 & \cellcolor{gray!10}\textbf{70.1} & \cellcolor{gray!30}\textbf{97.7} & 88.1 \\
DynamicVis-L & \cellcolor{gray!70}\textbf{71.5} & \cellcolor{gray!70}\textbf{97.8} & 85.2 & \cellcolor{gray!70}\textbf{71.0} & \cellcolor{gray!70}\textbf{97.8} & \cellcolor{gray!30}\textbf{89.2} \\
\bottomrule
\end{tabular}
}
\end{minipage}
\hfill
\begin{minipage}[t]{0.299\linewidth}
\centering
\caption{
mAP@20 metrics for zero-shot image retrieval on BigEarthNet (BE) and ForestNet (FN) benchmarks.} \label{tab:sota_retrival}
\resizebox{1\linewidth}{!}{
\begin{tabular}{l c l c c c c}
\toprule
Model & Band & Method & BE-43 & BE-19 & FN-12 & FN-4 \\
\midrule
\multirow{3}{*}{Prithvi \cite{jakubik2023foundation}} & \multirow{3}{*}{all} & Embedding & 97.62 & 97.98 & 44.51 & 60.76 \\
& & Binary emb. & 97.44 & 97.83 & 43.28 & 59.85  \\
& & 64-bit hash & 92.58 & 93.44 & 41.49 & 55.93  \\
\midrule
\multirow{3}{*}{SatMAE \cite{cong2022satmae}} & \multirow{3}{*}{all} & Embedding & 94.78 & 95.59 & 37.61 & 52.94  \\
& & Binary emb. & 89.39 & 90.40 & 36.49 & 53.04  \\
& & 64-bit hash & 79.35 & 80.99 & 30.79 & 47.19 \\
\midrule
\midrule
\multirow{3}{*}{Prithvi \cite{jakubik2023foundation}}& \multirow{3}{*}{RGB} & Embedding & 92.15 & 93.17 & 38.65 & 53.85 \\
& & Binary emb. & 91.38 & 92.43 & 38.11 & 53.31\\
& & 64-bit hash & 82.60 & 84.45 & 32.58 & 48.20\\
\midrule
\multirow{3}{*}{ViT-B \cite{dosovitskiy2020image}} & \multirow{3}{*}{RGB} & Embedding & 89.31 & 90.21 & 38.92 & 56.49 \\
&  &Binary emb. & 88.71 & 89.70 & 39.19 & 57.01 \\
&  &64-bit hash & 79.01 & 81.54 & 33.60 & 49.63 \\
\midrule
\midrule
\multirow{3}{*}{DynamicVis-B} & \multirow{3}{*}{RGB} & Embedding & 94.65 & 95.59 & 44.15 & 62.92 \\
& &Binary emb. & 94.07 & 95.04 & 43.94 & 62.64 \\
& &64-bit hash & 88.66 & 89.95 & 36.50 & 56.31 \\
\midrule
\multirow{3}{*}{DynamicVis-L} & \multirow{3}{*}{RGB} & Embedding & 94.98 & 95.82& 44.86&  63.59 \\
& &Binary emb. & 94.28 & 95.46 & 44.71& 63.28 \\
& &64-bit hash  & 89.15 & 90.58& 36.98& 56.86 \\
\bottomrule
\end{tabular}
}
\end{minipage}
\end{table*}

\begin{figure}[!htbp]
\centering
\includegraphics[width=\linewidth]{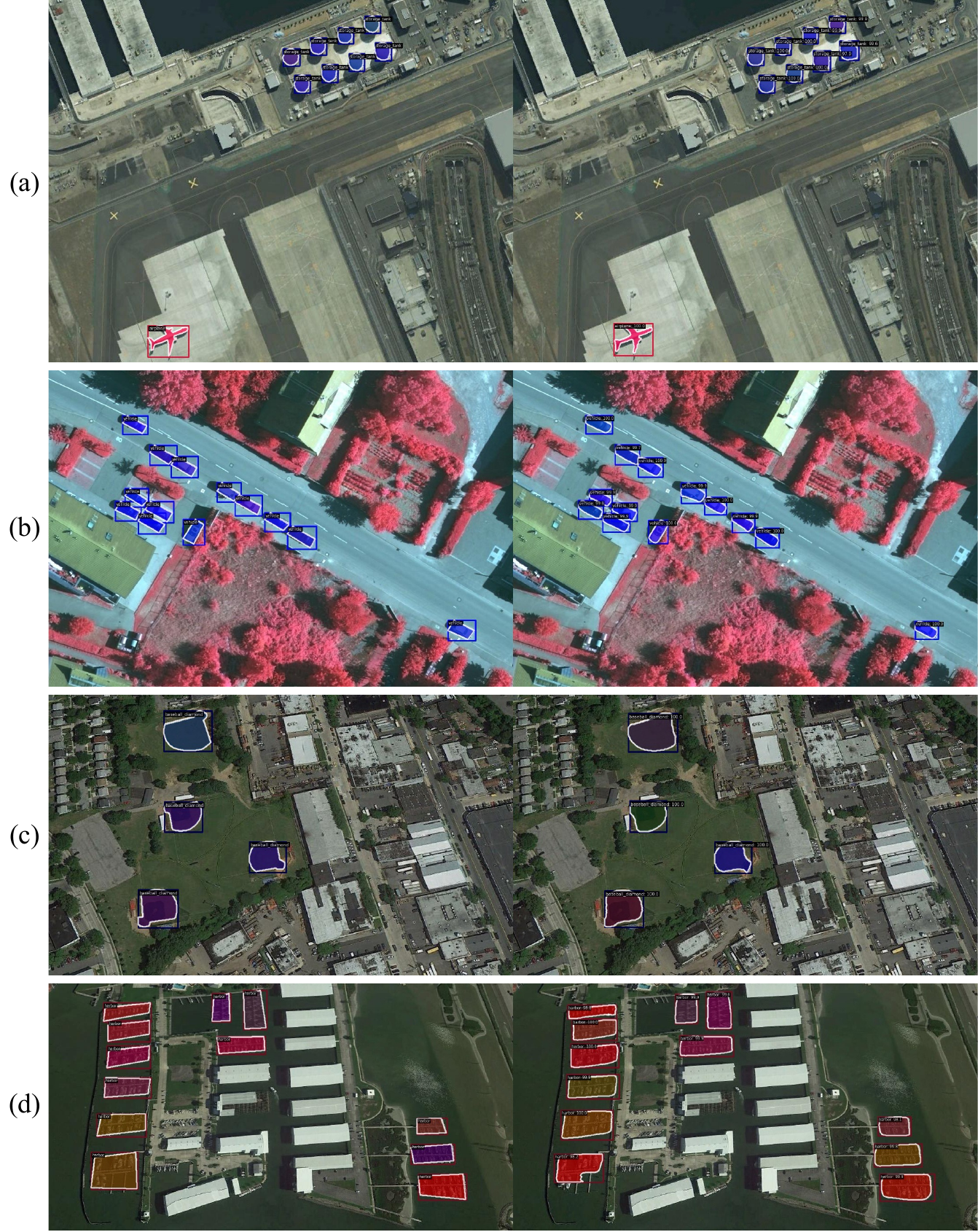}
\caption{
Qualitative instance segmentation results of DynamicVis-L on the NWPU VHR-10 test set. Left: Ground-truth annotations; Right: Model predictions. Note the exceptional preservation of boundaries in densely structured scenes.
} \label{fig:vis_nwpu}
\end{figure}

\begin{figure}[!htbp]
\centering
\includegraphics[width=\linewidth]{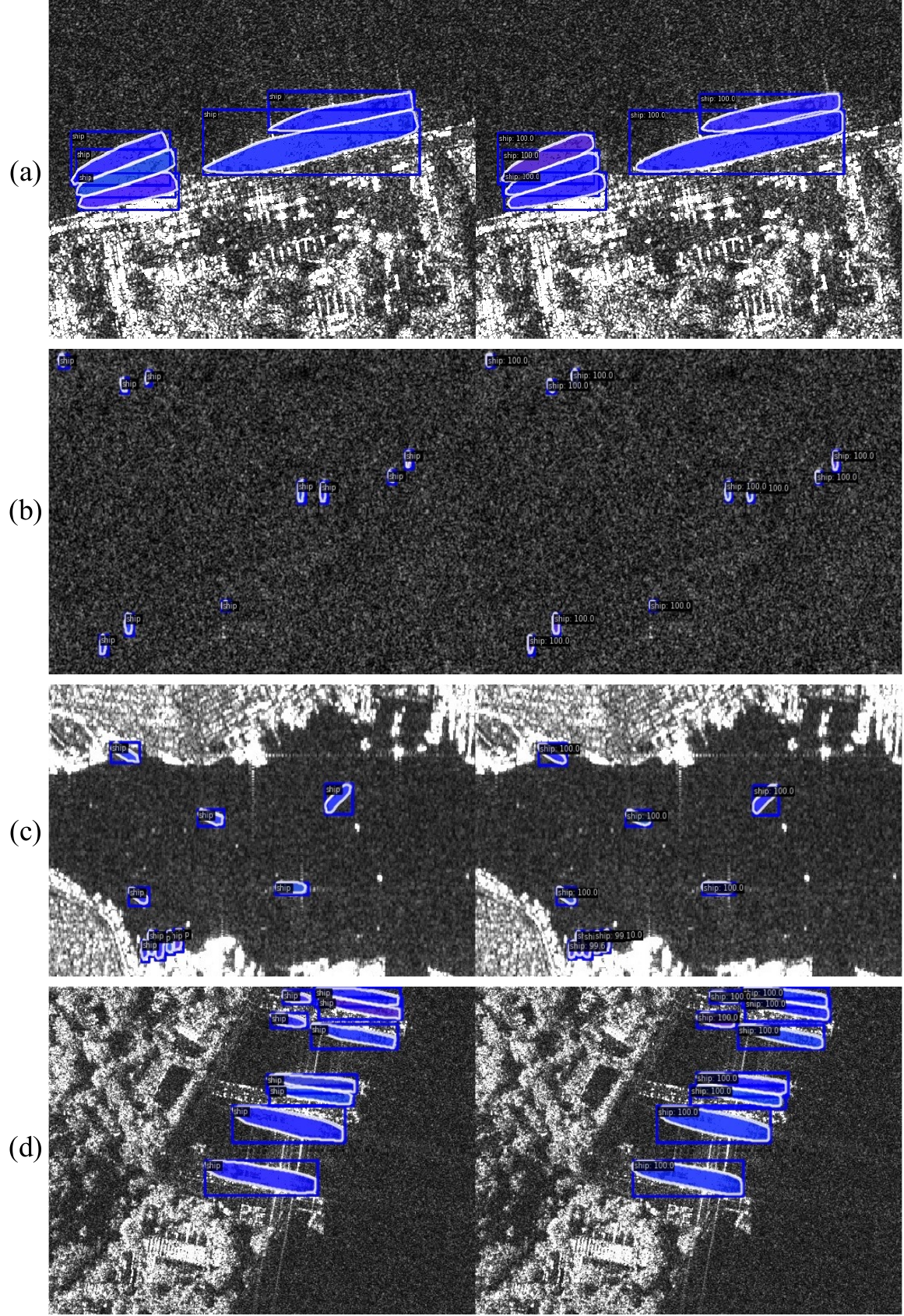}
\caption{
Qualitative segmentation results generated by DynamicVis-L on the SAR-based SSDD test set. Left: Ground-truth annotations; Right: Model predictions. The model easily adapts to extreme modality-specific noise.
} \label{fig:vis_ssdd}
\end{figure}

\textbf{Visualizations:} As demonstrated in Figs. \ref{fig:vis_nwpu} and \ref{fig:vis_ssdd}, DynamicVis-L executes robust instance segmentation across challenging scenarios. The model exhibits exceptional edge preservation and boundary crispness for artificially structured targets, effortlessly disentangling densely clustered objects while avoiding the fragmentation typical of aggressive patch-dropping transformer variants.


\subsection{Evaluation on Pixel-Level Dense Prediction}

While aggressive token-dropping techniques (e.g., DynamicViT) often incur catastrophic failures on dense prediction tasks due to disrupted spatial topologies \cite{rao2021dynamicvit}, DynamicVis elegantly circumvents this limitation. By leveraging a Context-Preserving Incremental Connection, the architecture retains unselected background tokens via parameter-free residual links, guaranteeing the macroscopic structural continuity required for pixel-level mapping.

\subsubsection{Bi-temporal Change Detection}

\textbf{Datasets \& Implementations:}
Change detection is inherently a sparse prediction task, as temporal discrepancies typically occupy a minimal fraction of bi-temporal image pairs. We comprehensively evaluate the proposed framework on three standard benchmarks encompassing diverse data sources, spatial scales, and resolutions: LEVIR-CD \cite{chen2020spatial} (0.5m Google Earth imagery), WHU-CD \cite{ji2018fully} (0.3m aerial imagery of disaster-affected regions), and OSCD \cite{daudt2018urban} (multi-resolution Sentinel-2 multispectral imagery). Bi-temporal image pairs are fed into the pre-trained DynamicVis backbone to extract hierarchical multi-scale features, which are then concatenated and processed by an MLP decoder. The network is optimized using a composite objective of Cross-Entropy and Dice loss via the AdamW optimizer for 300 epochs. Input images are standardized to $512 \times 512$ or $256 \times 256$ patches, with a $2\times$ variant utilizing bilinear upsampling to double spatial dimensions and preserve fine-grained structural details.

\textbf{Results and Analysis:}
Quantitative evaluations across the three benchmarks (Tables \ref{tab:levir-cd}, \ref{tab:whu-cd}, and \ref{tab:oscd}) demonstrate that DynamicVis consistently achieves state-of-the-art performance. Three critical insights emerge from the analysis:
\textit{i) Parameter Efficiency vs. Massive Foundation Models:} Despite possessing only millions of parameters, DynamicVis matches or strictly exceeds the performance of massive foundation models containing billions of parameters (e.g., SkySense \cite{guo2024skysense}, MTP-IMP \cite{wang2024mtp}) and large-scale MAE-based architectures (e.g., SpectralGPT \cite{hong2024spectralgpt}, RingMo \cite{sun2022ringmo}).
\textit{ii) Superiority in Extreme Sparsity:} On the OSCD dataset (Table \ref{tab:oscd}), which is characterized by extreme change sparsity, DynamicVis establishes a profound advantage, achieving an F1 score of 60.25\%. This confirms the efficacy of our inductive bias: by explicitly routing computational resources to salient structural tokens, the model easily highlights true temporal discrepancies while actively ignoring trivial background noise, such as seasonal illumination shifts.
\textit{iii) Resolution and Pre-training Scaling:} Increasing the resolution (DynamicVis-L$^{2\times}$) and applying Meta-Embedding MIL pre-training protocol consistently yield substantial performance gains, proving the framework's adaptability in modeling complex spatial-spectral variations commonly found in high-resolution datasets like WHU-CD.

\begin{figure*}[t]
\centering
\includegraphics[width=\linewidth]{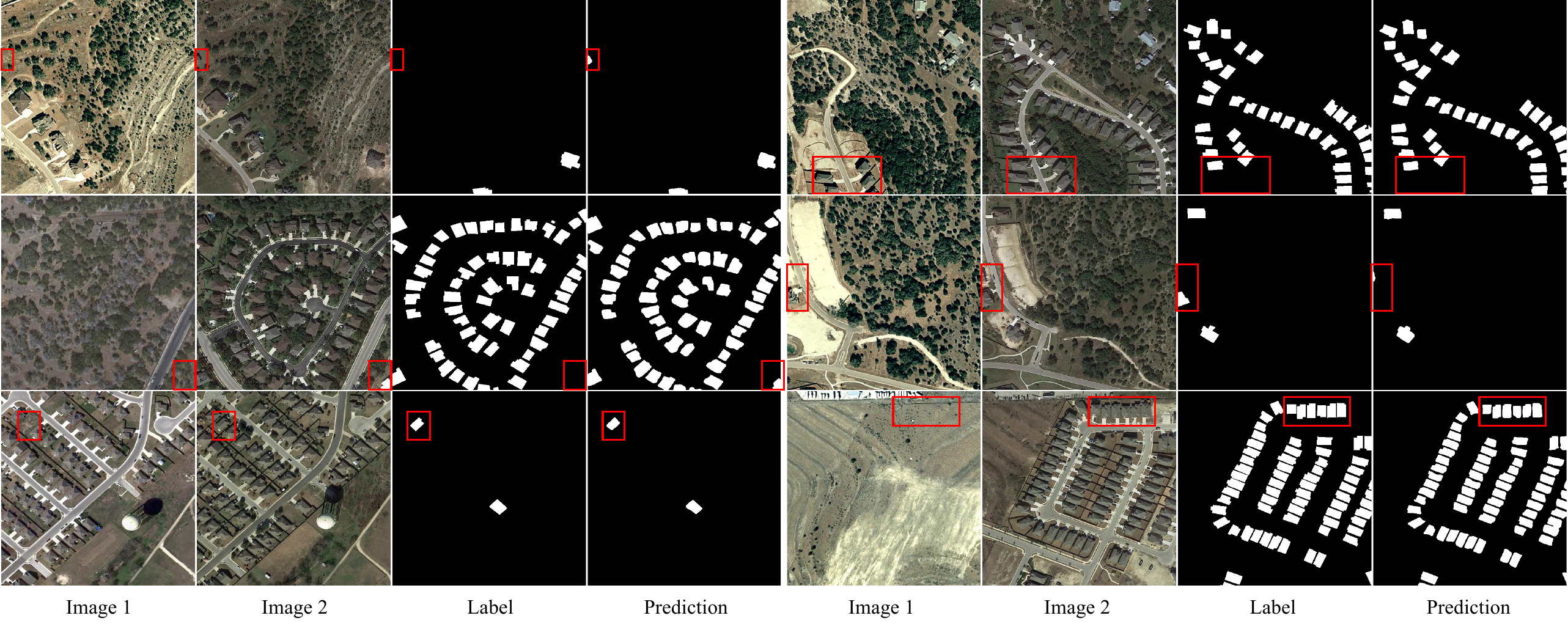}
\caption{Detection samples from DynamicVis on the LEVIR-CD test set. The \textcolor{red}{\textbf{red}} boxes highlight the capability of DynamicVis to accurately depict ``changes" in complex scenes, correctly delineating instances that are severely obfuscated or have been overlooked by human annotators.}
\label{fig:vis_levircd}
\end{figure*}

\textbf{Visualizations:}
Qualitative assessments on the LEVIR-CD test set (Fig. \ref{fig:vis_levircd}) confirm that DynamicVis exhibits exceptional edge preservation and boundary crispness. The model effortlessly disentangles densely clustered changes and identifies subtle temporal variations overlooked even by human annotators, while maintaining robust resistance against pseudo-changes caused by lighting or seasonal variations.

\begin{table*}[!tbhp]
\centering
\begin{minipage}[t]{0.35\linewidth}
\centering
\caption{
Comparisons of various change detection methods evaluated on the LEVIR-CD test dataset.
} \label{tab:levir-cd}
\resizebox{\linewidth}{!}{
\begin{tabular}{l c c c c}
\toprule
Method & P & R & F1 & IoU \\
\midrule
FC-EF \cite{daudt2018fully} & 86.91& 80.17 &83.40 &71.53 \\
FC-Siam-Diff \cite{daudt2018fully} & 89.53 &83.31 &86.31 &75.91 \\
SNUNet \cite{fang2021snunet} &89.18 &87.17 &88.16 &78.83 \\
BIT \cite{chen2021remote} & 89.24 &89.37 &89.30 &80.68 \\
ChangeStar \cite{zheng2021change} & - & - & 91.25 & 83.92 \\
Changen \cite{zheng2023scalable} & - & - & 91.50 & - \\
HCGMNet \cite{han2023hcgmnet} & 92.96 & 90.61 & 91.77 & 84.79 \\
ChangerEx \cite{fang2023changer} & 92.97 & 90.61 & 91.77 & - \\
WNet \cite{tang2023wnet}  & 91.16 & 90.18 & 90.67 & 82.93 \\
C2FNet \cite{han2024c2f} & \cellcolor{gray!10}\textbf{93.69} & 90.04 & 91.83 & 84.89 \\
P2V-CD \cite{lin2022transition} & 93.32 & 90.60 & 91.94 & - \\
ChangeCLIP-ViTB \cite{dong2024changeclip} & 93.68 & 89.04 & 91.30 & 83.99 \\
BAN-BIT \cite{li2024new} & 92.83 & \cellcolor{gray!30}\textbf{90.89} & 91.85 & \cellcolor{gray!10}\textbf{84.93} \\
GFM \cite{mendieta2023towards}  & - & - & 91.73 & - \\
SatLas \cite{bastani2023satlaspretrain} & - & - & 90.62 & - \\
CACo \cite{mall2023change} & - & - & 81.04 & - \\
SatMAE \cite{cong2022satmae} & - & - & 87.65 & - \\
RVSA \cite{wang2022advancing} & - & - & 90.86 & - \\
ChangeFormer \cite{bandara2022transformer} & 92.05 &88.80& 90.40 &82.47 \\
ICIFNet \cite{feng2022icif}  & -& -& 89.96 & 81.75 \\
DMINet \cite{feng2023change} & -& -& 90.71 & 82.99 \\
GASNet \cite{zhang2023global} & -& -& 90.52 & 83.48 \\
DDPM-CD \cite{bandara2025ddpm} & - & - & 90.91 & 83.35 \\
AMTNet \cite{liu2023attention} & 91.82 &89.71 &90.76 &83.08 \\
BiFA \cite{zhang2024bifa} & 91.52 & 89.86 & 90.69 & 82.96 \\
RSM-CD \cite{zhao2024rs} & 92.52 & 89.73 & 91.10 & 83.66 \\
CDMamba \cite{zhang2025cdmamba} & 91.43 & 90.08 & 90.75 & 83.07 \\
ChangeMamba \cite{chen2024changemamba} & 91.59 & 88.78 & 90.16 & 82.09 \\
SeCo-BiT-R50 \cite{manas2021seasonal} & - & - & 90.14 & - \\
RSP-BIT-VITAEv2-S \cite{wang2022advancing} & - & - & 90.93 & - \\
ChangeViT-T \cite{zhu2025changevit} & -& -& 91.81& 84.86 \\
RingMo-BIT-SwinB \cite{sun2022ringmo} & 92.47 & \cellcolor{gray!70}\textbf{91.17} & 91.85 & - \\
Scale-MAE  \cite{reed2023scale} & - & - & 92.07 & - \\
MTP-IMP \cite{wang2024mtp} & - & - & \cellcolor{gray!30}\textbf{92.54} & - \\
SkySense \cite{guo2024skysense} & - & - & \cellcolor{gray!70}\textbf{92.58} & - \\
\midrule
DynamicVis-B$^\dag$  & 93.22 & 89.54 & 91.34& 84.06 \\
DynamicVis-B$^\ddag$ & 92.89 &  89.02 & 90.92 & 83.32 \\
DynamicVis-B  & 93.65 & 90.09 & 91.82 & 84.75 \\
DynamicVis-B$^{2\times}$ & \cellcolor{gray!30}\textbf{93.79} & 90.36& 92.05& \cellcolor{gray!30}\textbf{85.19} \\
DynamicVis-L$^\dag$ & 92.73 & \cellcolor{gray!10}\textbf{90.69} & 91.69 & 84.66 \\
DynamicVis-L$^\ddag$  & 93.00 & 89.51 & 91.22 & 83.86 \\
DynamicVis-L & 93.52 &  90.15 & 91.90 & 84.81 \\
DynamicVis-L$^{2\times}$ & \cellcolor{gray!70}\textbf{93.97} &  90.48 & \cellcolor{gray!10}\textbf{92.32} & \cellcolor{gray!70}\textbf{85.31} \\
\bottomrule
\end{tabular}}
\end{minipage}
\hfill
\begin{minipage}[t]{0.63\linewidth}
\centering
\begin{minipage}[t]{0.49\linewidth}
\centering
\caption{
Comparisons of various change detection methods evaluated on the WHU-CD test dataset.
} \label{tab:whu-cd}
\resizebox{\linewidth}{!}{
\begin{tabular}{l c c c c}
\toprule
Method & P & R & F1 & IoU \\
\midrule
FC-EF \cite{daudt2018fully} & 78.86 & 78.64 & 78.75 & 64.95 \\
FC-Siam-Diff \cite{daudt2018fully} & 84.73 & 87.31 & 86.00  &75.44 \\
STANet \cite{chen2020spatial} & 79.37 & 85.50  &82.32 & 69.95 \\
SNUNet \cite{fang2021snunet} & 85.60  &81.49  &83.49 & 71.67 \\
BIT \cite{chen2021remote} & 86.64  &81.48  &83.98  &72.39 \\
ChangeFormer \cite{bandara2022transformer} & 90.09 & 84.85 & 87.39 &77.61 \\
GCD-DDPM \cite{wen2024gcd} & 92.79 & 92.29 & 92.54 & 86.52 \\
CGNet \cite{han2023change} & 94.47 & 90.79 & 92.59 & 86.21  \\
WNet \cite{tang2023wnet} & 92.37 & 90.15 & 91.25 & 83.91 \\
P2V-CD \cite{lin2022transition} & 95.48 & 89.47 & 92.38 & - \\
DDPM-CD \cite{bandara2025ddpm} & - & - & 92.65 & 86.31 \\
FresUNet \cite{daudt2019multitask}  & 86.55 & 77.68 & 81.88 & 69.32 \\
ICIFNet \cite{feng2022icif} & 92.98 & 85.56 & 88.32 & 79.24 \\
DMINet \cite{feng2023change} & 93.84 & 86.25 &88.69 &79.68 \\
GASNet \cite{zhang2023global} & -& -&91.75 &84.76 \\
EATDer \cite{ma2023eatder}  &91.32 & 88.74 &90.01 &81.97 \\
MTCNet \cite{shu2022mtcnet}  & 75.10  & \cellcolor{gray!30}\textbf{91.90}  &82.65  &70.43 \\
MSCANet \cite{dong2021multiscale} & 91.10  &89.86  &90.47  &82.60 \\
RSM-CD \cite{zhao2024rs} & 93.37 & 90.42 & 91.87 & 84.96 \\
ChangeMamba \cite{chen2024changemamba} & 94.21 & 90.94 & 92.55 & 86.13 \\
CDMamba \cite{zhang2025cdmamba} & 95.58 & 92.01 & 93.76 & \cellcolor{gray!10}\textbf{88.26} \\
PA-Former \cite{liu2022pa} & 94.28 & 90.38 & 92.29 & 85.69 \\
DARNet \cite{li2022densely} & 91.99 & 91.17 & 91.58 & 84.46 \\
MTP-IMP \cite{wang2024mtp} & - & - & \cellcolor{gray!70}\textbf{95.59} & - \\
\midrule
DynamicVis-B$^\dag$  & 96.43 & 89.83 &   93.01 & 86.94 \\
DynamicVis-B$^\ddag$  & 95.63 & 89.23 &   92.61 & 86.04 \\
DynamicVis-B & \cellcolor{gray!10}\textbf{96.56} & 91.48 & 93.73 & 88.06 \\
DynamicVis-B$^{2\times}$ & \cellcolor{gray!70}\textbf{96.82} & \cellcolor{gray!10}\textbf{91.86} & \cellcolor{gray!10}\textbf{94.27} & \cellcolor{gray!30}\textbf{89.52} \\
DynamicVis-L$^\dag$  & 96.41 & 90.04 & 93.12 & 87.11 \\
DynamicVis-L$^\ddag$  & 96.01 & 89.64 & 92.71 & 86.91 \\
DynamicVis-L &  96.41 & 90.83 & 93.60 & 87.91 \\
DynamicVis-L$^{2\times}$ & \cellcolor{gray!30}\textbf{96.78} & \cellcolor{gray!10}\textbf{92.50} &  \cellcolor{gray!30}\textbf{94.79} & \cellcolor{gray!30}\textbf{89.85} \\
\bottomrule
\end{tabular}}
\end{minipage}
\hfill
\begin{minipage}[t]{0.47\linewidth}
\centering
\caption{
Comparisons of various change detection methods evaluated on the OSCD test dataset.
} \label{tab:oscd}
\resizebox{1\linewidth}{!}{
\begin{tabular}{l c c c c}
\toprule
Method & P & R & F1 & IoU \\
\midrule
SNUNet \cite{fang2021snunet} & - & - & 27.02 & 15.62  \\
ChangeFormer \cite{bandara2022transformer} & - & - &38.22 & 23.62 \\
BIT \cite{chen2021remote} & - & - &29.58 & 17.36 \\
SwiMDiff \cite{tian2024swimdiff} & 63.60 & 40.90 & 49.60 & - \\
GASNet \cite{zhang2023global} & - & - &10.71 & 5.66  \\
AMTNet \cite{liu2023attention} & - & - &10.25 & 5.40 \\
EATDer \cite{ma2023eatder} & - & - &54.23 & 36.98  \\
ChangeViT-T \cite{zhu2025changevit} & - & - &55.13 & 38.06  \\
ChangeViT-S \cite{zhu2025changevit} & - & - &55.51 & 38.42 \\
GASSL \cite{ayush2021geography} & - & - & 46.26 & - \\
SeCo \cite{manas2021seasonal} & 57.71 & 49.23 & 49.82 & - \\
CACo \cite{mall2023change} & 62.87 & 44.49 & 52.11 & - \\
MoCo-v2 \cite{chen2020improved} & 64.49 & 30.94 &40.71 & - \\
Swin-22k \cite{liu2021swin} & 46.88 & 59.28 & 52.35 & - \\
ViT-22k \cite{dosovitskiy2020image} & 52.09 & 52.37 & 52.23 & - \\
SatMAE \cite{cong2022satmae} & 55.18 & 50.54 & 52.76 & - \\
DINO-MC \cite{wanyan2023dino} & - & - & 52.70 & - \\
MTP-IMP \cite{wang2024mtp} & - & - & 55.61 & - \\
SpectralGPT \cite{hong2024spectralgpt} & 51.65 & 56.15 & 53.51 & - \\
SpectralGPT$^+$ \cite{hong2024spectralgpt} & 52.39 & \cellcolor{gray!30}\textbf{57.20} & 54.29 & - \\
MATTER \cite{akiva2022self} & 61.80 & \cellcolor{gray!10}\textbf{57.13} & 59.37 & - \\
GFM \cite{mendieta2023towards} & 58.07 & \cellcolor{gray!70}\textbf{61.67} & 59.82 & - \\
SkySense \cite{guo2024skysense} & - & - & \cellcolor{gray!30}\textbf{60.06} & - \\
\midrule
DynamicVis-B$^\dag$  & \cellcolor{gray!10}\textbf{75.08} & 44.37 & 55.78 & 38.67 \\
DynamicVis-B$^\ddag$  & 71.21 & 41.03 & 52.06 & 35.19 \\
DynamicVis-B & 74.28 & 45.24 & 56.23 & 39.11 \\
DynamicVis-B$^{2\times}$ & \cellcolor{gray!30}\textbf{77.52} & 48.81 & \cellcolor{gray!10}\textbf{59.90} &  \cellcolor{gray!30}\textbf{42.75} \\
DynamicVis-L$^\dag$  & 64.83 & 49.20 & 55.95 & 38.84 \\
DynamicVis-L$^\ddag$  & 74.42 & 42.23 & 53.88 & 36.88 \\
DynamicVis-L &  72.77 & 46.99 & 57.11 & \cellcolor{gray!10}\textbf{39.96} \\
DynamicVis-L$^{2\times}$ & \cellcolor{gray!70}\textbf{79.41} & 48.36 & \cellcolor{gray!70}\textbf{60.25} & \cellcolor{gray!70}\textbf{43.16} \\
\bottomrule
\end{tabular}
}
\end{minipage}

\begin{minipage}[t]{0.47\linewidth}
\vspace{-0.02cm}
\centering
\caption{Region classification performance on the fMoW test set.
} \label{tab:fmow}
\resizebox{1\linewidth}{!}{
\begin{tabular}{l c c c}
\toprule
Method & P & R & F1 \\
\midrule
ResNet-50 \cite{he2016deep} & 95.86 & 95.61 & 95.56 \\
ViT-B \cite{dosovitskiy2020image} & 96.62 & 96.87 & 96.68 \\
\midrule
DynamicVis-B (CE) & 96.45 & 96.54 & 96.37 \\
DynamicVis-B (MIL) & 97.49 & 97.03 & 97.19 \\
DynamicVis-L (CE) & 97.05 & 96.72 & 96.84  \\
DynamicVis-L (MIL) & \cellcolor{gray!50}\textbf{97.91} & \cellcolor{gray!50}\textbf{97.97} & \cellcolor{gray!50}\textbf{97.87}  \\
\bottomrule
\end{tabular}}
\end{minipage}
\hfill
\begin{minipage}[t]{0.47\linewidth}
\vspace{-0.45cm}
\centering
\caption{Inference efficiency across various models with $512 \times 512$ input.
} \label{tab:ab_backbone_efficiency}
\resizebox{\linewidth}{!}{
\begin{tabular}{l c c c c}
\toprule
\multirow{2}{*}{Model} & Max & Params.  & FLOPs  & Throughput \\
& BS & (M) & (G) & (Sampels/s) \\
\midrule
ResNet18 \cite{he2016deep}& 1200 & 11.69 & 9.50 & 1200 \\
ResNet50 \cite{he2016deep}& 642 & 25.56 & 21.47& 340 \\
ResNet101 \cite{he2016deep}& 608 & 44.55 &40.92 & 200 \\
ViT-B \cite{dosovitskiy2020image}& 268 & 87.20 & 87.76 & 86 \\
ViT-L \cite{dosovitskiy2020image}& 208 & 305.00 & 362.00 & 26 \\
\midrule
DynamicVis-B$^\dag$& 186 & 36.76 & 54.28 & 90  \\
DynamicVis-B & 998 & 36.77 & 30.07 & 196  \\
DynamicVis-L$^\dag$ & 132 & 91.27 & 151.00 & 45 \\
DynamicVis-L & 786 & 91.29 & 82.31 & 92 \\
\bottomrule
\end{tabular}}
\end{minipage}
\end{minipage}
\end{table*}

\subsubsection{Semantic Segmentation}

\textbf{Datasets \& Implementations:}
We evaluate semantic segmentation performance on two single-category benchmarks with fundamentally contrasting morphological properties: the Massachusetts roads dataset \cite{mnih2013machine} (characterized by continuous, elongated linear topologies) and the WHU building dataset \cite{ji2018fully} (comprising discrete, localized instance-like structures). We adopt UperNet \cite{xiao2018unified} as the dense prediction framework, initializing both the backbone and FPN neck with pre-trained DynamicVis weights.

\textbf{Results and Analysis:}
As detailed in Tables \ref{tab:massachusetts_sota} and \ref{tab:whu_building_sota}, DynamicVis-L$^{2\times}$ attains highly competitive F1 scores of 80.35\% on Roads and 95.58\% on Buildings. A nuanced dichotomy in performance emerges based on geometric topology:
\textit{i) Discrete vs. Continuous Inductive Biases:} The model exhibits absolute superiority on the WHU dataset, as discrete building instances perfectly align with the framework's instance-aware dynamic routing mechanism. Conversely, segmenting continuous road poses a unique challenge for localized token modeling; hence, DynamicVis slightly surpasses heavily specialized, geometry-constrained architectures (e.g., GA-Net \cite{chen2022ga}). Crucially, compared to standard token-pruning models that completely fracture and fail on continuous structures, DynamicVis maintains excellent spatial connectivity, confirming the robustness of its parameter-free incremental residual connections.
\textit{ii) Efficacy of Meta-Embedding Pre-training:} The integration of geographic priors via MIL pre-training protocol yields substantial performance leaps compared to training from scratch, verifying that semantically rich token activation is necessary to prevent feature degradation in dense pixel-level mapping.

\begin{figure*}[!h]
\centering
\includegraphics[width=\linewidth]{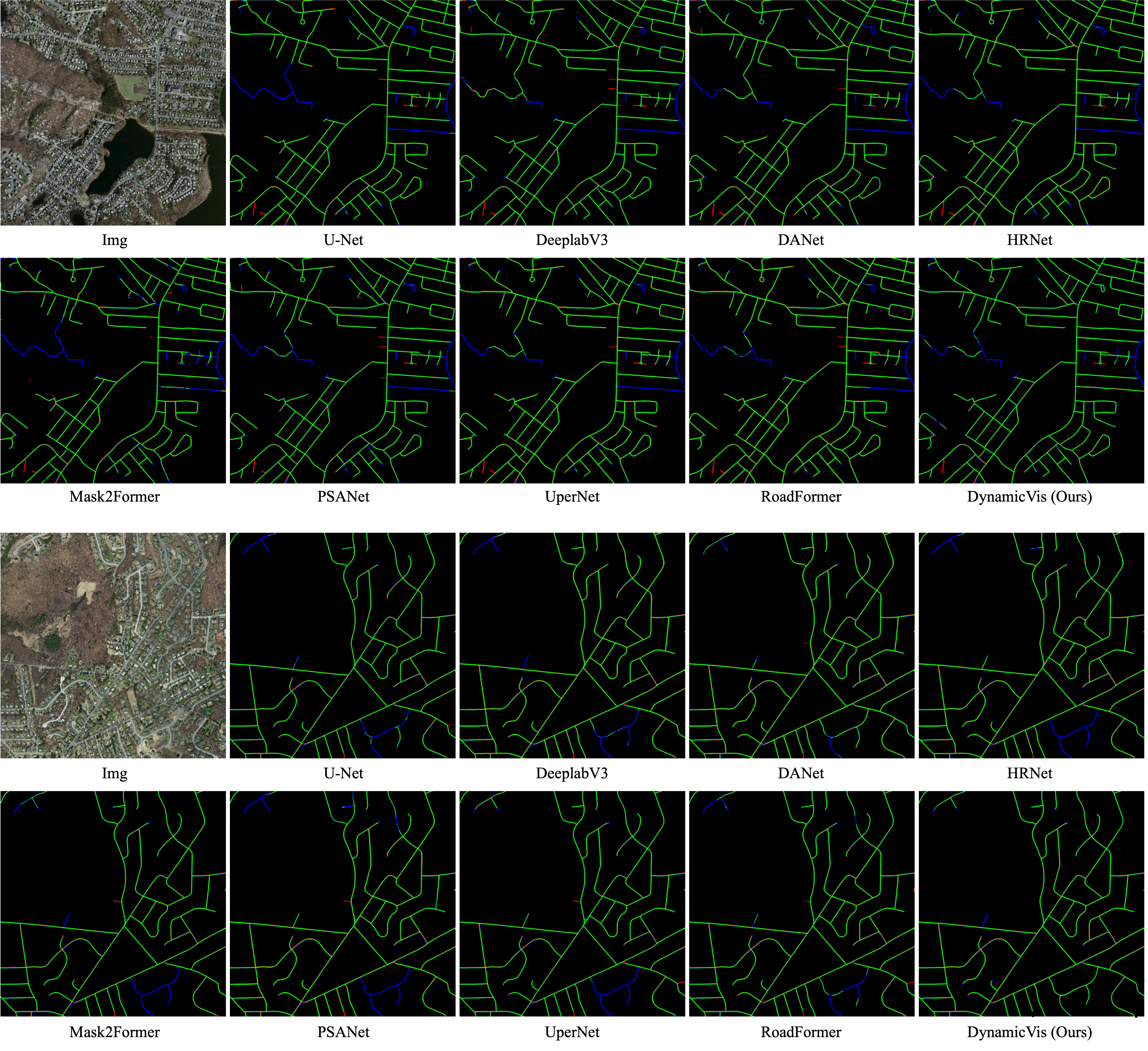}
\caption{Comparative analysis of road detection performance on the Massachusetts road dataset. True positive (TP), false negative (FN), and false positive (FP) detections are annotated with \textcolor{green}{\textbf{green}}, \textcolor{blue}{\textbf{blue}}, and \textcolor{red}{\textbf{red}} markers, respectively.}
\label{fig:vis_massachusetts}
\end{figure*}

\begin{figure*}[t]
\centering
\includegraphics[width=\linewidth]{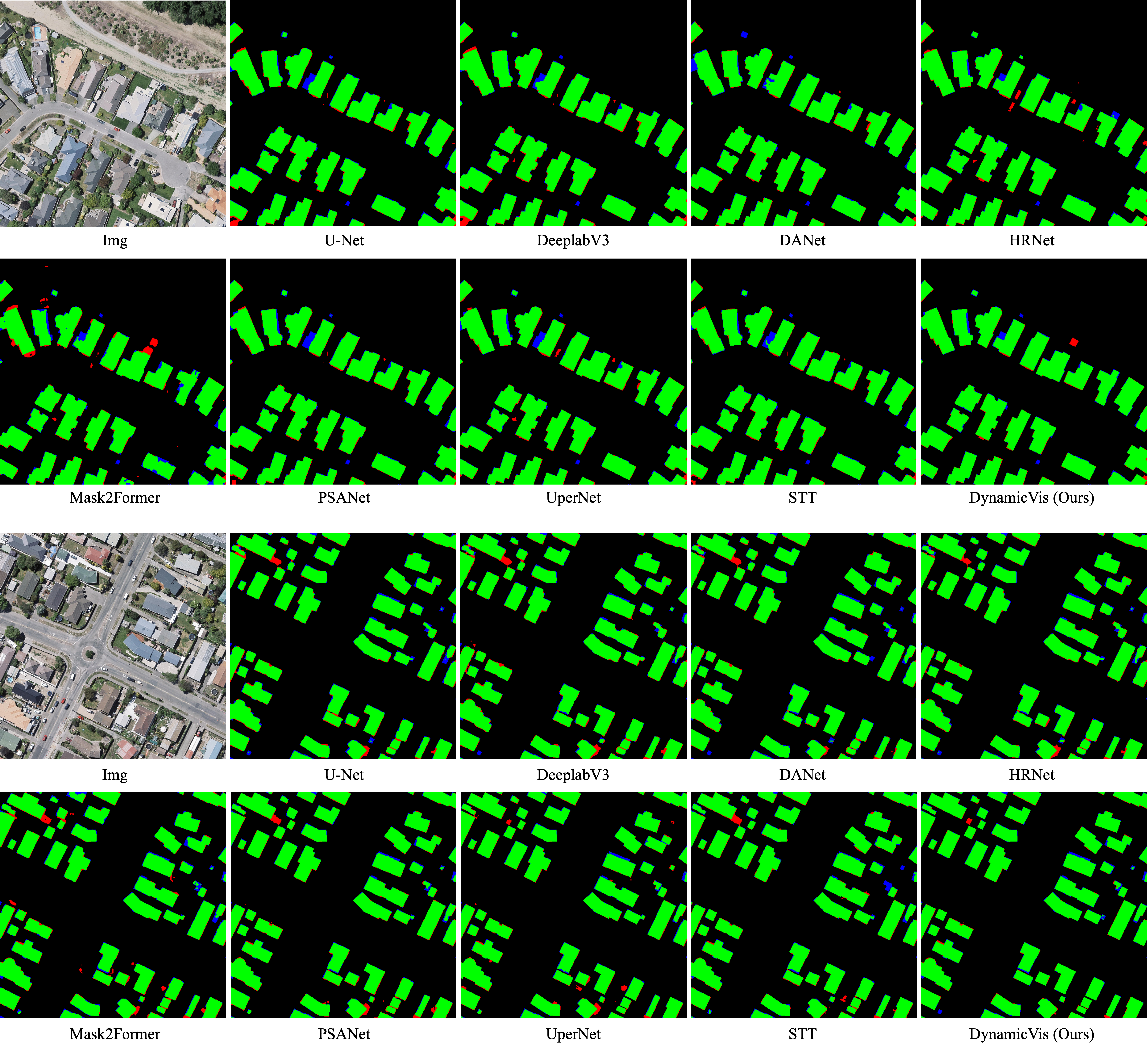}
\caption{Comparative analysis of building segmentation performance on the WHU building dataset. True positive (TP), false negative (FN), and false positive (FP) detections are annotated with \textcolor{green}{\textbf{green}}, \textcolor{blue}{\textbf{blue}}, and \textcolor{red}{\textbf{red}} markers, respectively.}
\label{fig:vis_whu_building}
\end{figure*}

\textbf{Visualizations:}
Qualitative comparisons in Fig. \ref{fig:vis_massachusetts} and Fig. \ref{fig:vis_whu_building} illustrate the efficacy of DynamicVis-L. The model successfully preserves complex structural continuities across diverse geometric configurations and demonstrates superior robustness under challenging conditions, such as severe occlusions by dense vegetation or low image clarity, where competing methods frequently hallucinate spurious detections or fail to localize critical features entirely.

\subsection{Evaluation on Region-Level Understanding}

\subsubsection{Scene and Region Classification}

\textbf{Datasets \& Implementations:}
To evaluate model's macroscopic semantic understanding, we conduct global scene classification on two diverse benchmarks: UC-Merced \cite{yang2011spatial} (21 categories, 0.3m resolution) and AID \cite{xia2017aid} (30 categories, 0.5--8m multi-resolution). Additionally, region-level classification is evaluated on the loosely annotated fMoW \cite{christie2018functional} test set. Standard data augmentation techniques were applied, and the pre-trained foundation models were fine-tuned via an MLP classification head.

\textbf{Results and Analysis:}
As detailed in Table \ref{tab:sota_uc_aid}, DynamicVis establishes a new state-of-the-art across both scene classification datasets. The large-scale variant (DynamicVis-L) approaches near-perfect accuracy on UC-Merced (99.12\% Precision) and decisively outperforms both transformer-based models (e.g., ViT \cite{dosovitskiy2020image}, Swin \cite{liu2021swin}) and recent Mamba-based architectures (e.g., Vim \cite{zhu2024vision}, VMamba \cite{liu2024vmamba}, RSMamba \cite{chen2024rsmamba}) on the more complex AID dataset.

Crucially, these results elucidate the inherent trade-offs of our spatial sparsity inductive bias. Because global scene classification often relies on holistic background context (e.g., leveraging the surrounding ocean to classify an island), aggressively routing sparse tokens introduces a marginal accuracy reduction compared to dense processing (DynamicVis-B$^\ddag$ vs. DynamicVis-B$^\dag$). However, this trade-off is elegantly resolved by our Region-Level Meta-Embedding MIL pre-training. By explicitly teaching the model to map visual features to language-aligned meta-embeddings, the MIL paradigm cultivates a highly separable latent space. Consequently, the fully equipped DynamicVis comprehensively recovers and exceeds the baseline accuracy while maintaining the massive computational acceleration afforded by dynamic routing. On the fMoW region test set (Table \ref{tab:fmow}), the MIL-trained variant significantly outperforms the standard Cross-Entropy counterpart (97.87\% vs. 96.84\% F1), confirming that robust semantic priors mitigate the ambiguity of localized classification.

\begin{figure}[!htbp]
\centering
\includegraphics[width=\linewidth]{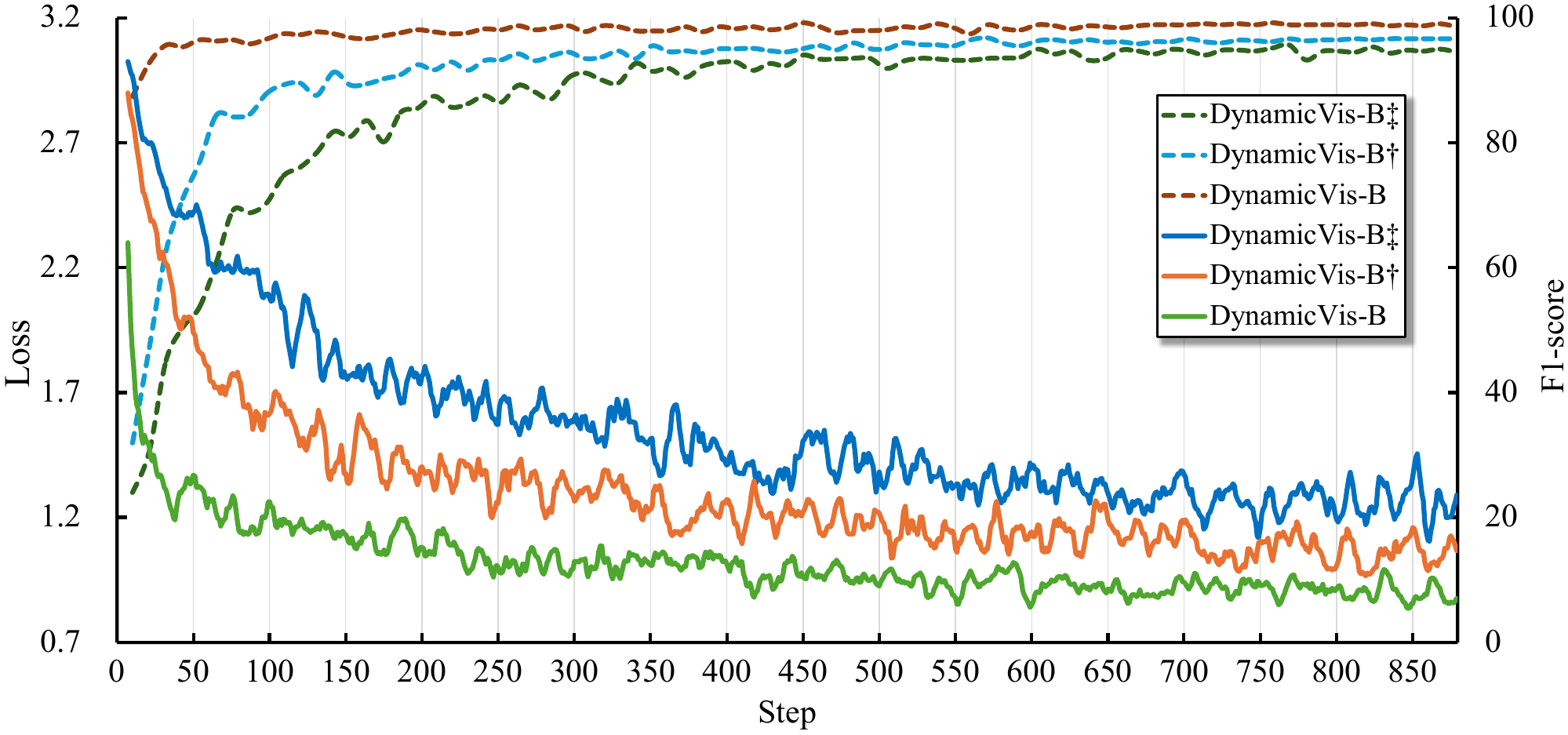}
\caption{Training loss (left y-axis) and evaluation accuracy (right y-axis) across iterations on the UC-Merced dataset. The inclusion of MIL pre-training ensures rapid convergence despite the optimization complexity of dynamic token routing.}
\label{fig:compare_uc_loss_f1}
\end{figure}

Furthermore, training dynamics on the UC-Merced dataset (Fig. \ref{fig:compare_uc_loss_f1}) illustrate that while the dynamic routing initially slows convergence as it learns to identify salient regions, the MIL pre-training drastically accelerates optimization, achieving peak accuracy in a fraction of the iterations required by standard dense architectures, underscoring the efficacy of meta-embeddings in guiding feature space convergence.

\begin{figure}[!htbp]
\centering
\includegraphics[width=\linewidth]{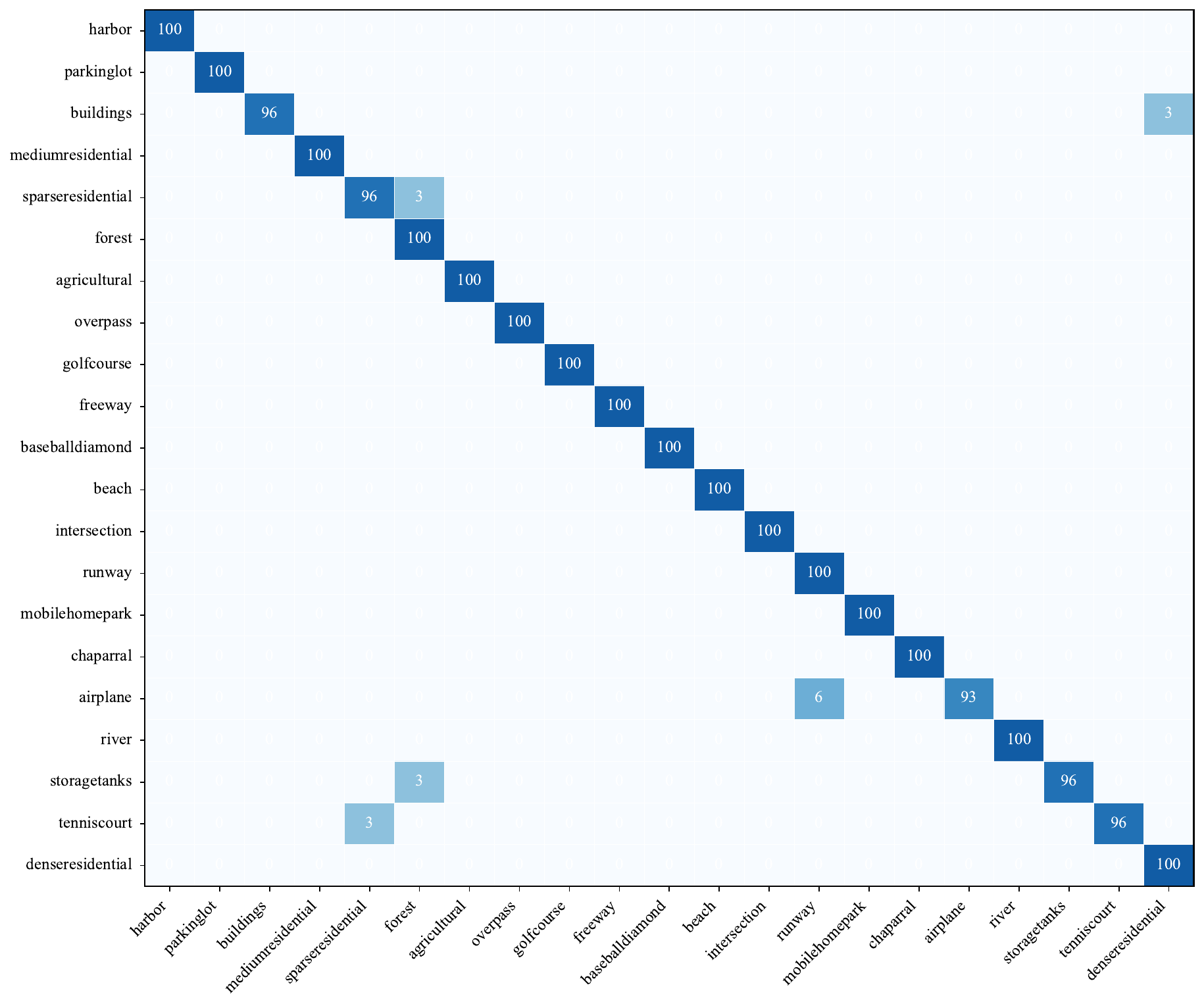}
\caption{Percentage-normalized confusion matrix for DynamicVis-L on the UC-Merced dataset. High diagonal values indicate precise scene classification, with minimal confusion restricted to highly correlated semantic pairs.}
\label{fig:uc_l_conf_mat}
\end{figure}

\begin{figure}[!htbp]
\centering
\includegraphics[width=\linewidth]{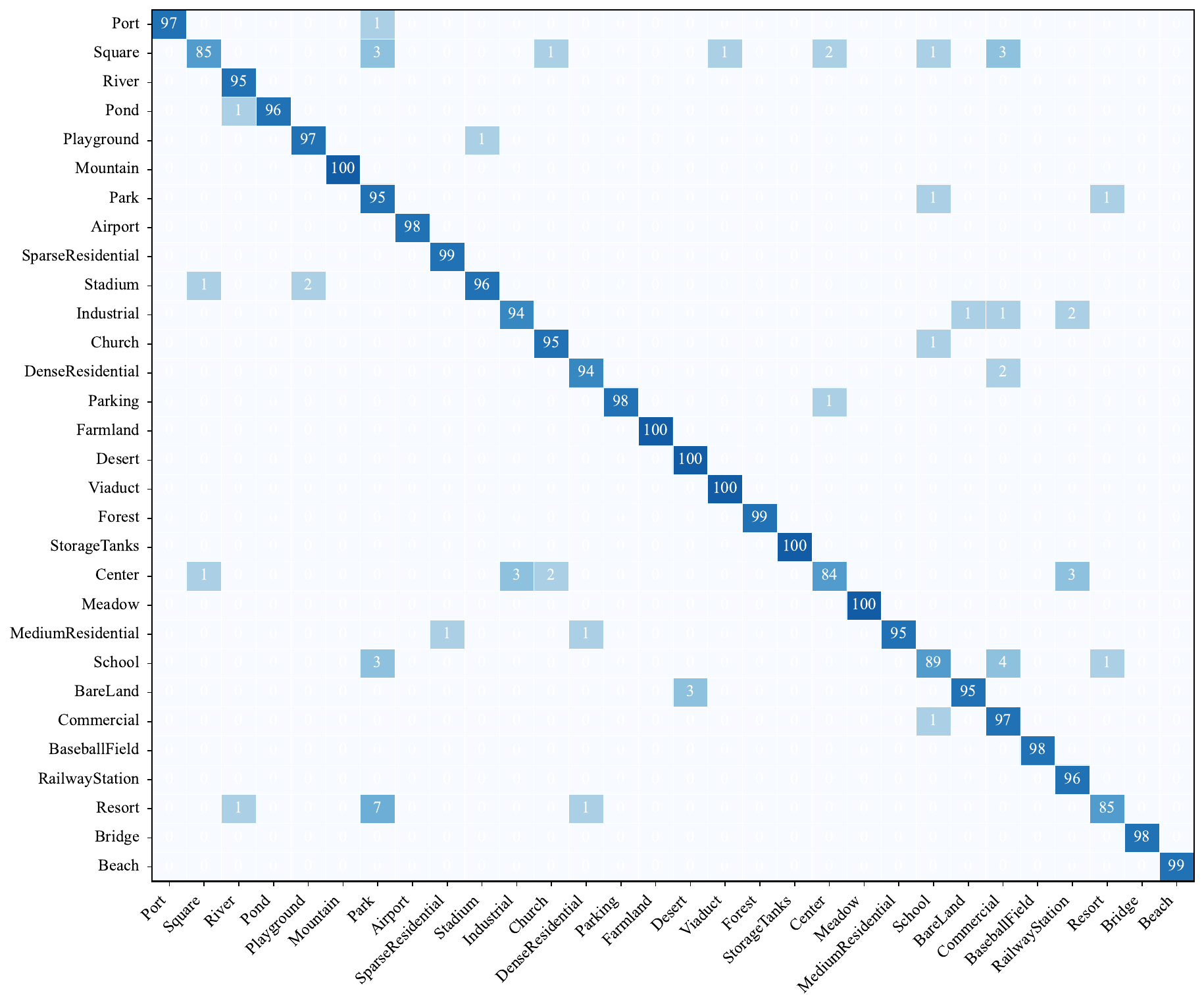}
\caption{Percentage-normalized confusion matrix for DynamicVis-L on the AID dataset. The model cleanly separates the expanded 30-class taxonomy, demonstrating robust macroscopic context encoding.}
\label{fig:aid_l_conf_mat}
\end{figure}

\textbf{Visualizations:}
The robustness of DynamicVis-L is further evidenced by its percentage-normalized confusion matrices (Fig. \ref{fig:uc_l_conf_mat} and Fig. \ref{fig:aid_l_conf_mat}). The near-perfect diagonal dominance reflects exceptional categorical discrimination. Misclassifications are rare and systematically localized to classes with profound spatial and semantic overlap. For instance, occasional confusion between ``airplane'' and ``runway'' in UC-Merced inherently stems from their strict spatial co-occurrence, while overlap between ``Resort'' and ``Park'' in AID reflects shared visual textures such as dense vegetation and recreational structures. Despite these challenging hard-negative scenarios, DynamicVis effectively isolates primary semantic targets.

\begin{table}[!tbp]
\centering
\caption{Comparative analysis with state-of-the-art methods across scene classification benchmarks. DynamicVis demonstrates robust superiority, particularly with the integration of its MIL pre-training paradigm.}
\label{tab:sota_uc_aid}
\setlength{\tabcolsep}{3pt}
\footnotesize
\begin{tabular}{l c| ccc ccc}
\toprule
\multirow{2}{*}{Method} & Params. & \multicolumn{3}{c}{UC Merced} & \multicolumn{3}{c}{AID} \\
\cmidrule(lr){3-5} \cmidrule(lr){6-8}
&(M) & P & R & F1 & P & R & F1 \\
\midrule
ResNet-18 \cite{he2016deep}& 11.7 & 87.98 & 87.46 & 87.40 & 88.70 & 88.17 & 88.30 \\
ResNet-50 \cite{he2016deep}& 25.6 & 91.99 & 91.74 & 91.65 & 89.44 & 88.66 & 88.87 \\
ResNet-101 \cite{he2016deep}& 44.6 & 92.40 & 92.22 & 92.12 & 91.03 & 90.63 & 90.81 \\
\midrule
DeiT-T \cite{touvron2021training}& 5.5 & 86.92 & 86.66 & 86.53 & 85.23 & 84.52 & 84.52 \\
DeiT-S \cite{touvron2021training}& 21.7 & 88.95 & 88.41 & 88.41 & 85.88 & 85.19 & 85.34 \\
DeiT-B \cite{touvron2021training}& 85.8 & 89.14 & 88.73 & 88.70 & 87.32 & 86.07 & 86.07 \\
ViT-B \cite{dosovitskiy2020image}& 87.2 & 91.09 & 90.79 & 90.77 & 89.39 & 88.65 & 88.86 \\
ViT-L \cite{dosovitskiy2020image}& 305.0 & 91.98 & 91.32 & 91.26 & 90.19 & 88.86 & 89.17 \\
Swin-T \cite{liu2021swin}& 27.5 & 90.87 & 90.63 & 90.40 & 86.49 & 85.66 & 85.77 \\
Swin-S \cite{liu2021swin}& 48.9 & 91.08 & 90.95 & 90.82 & 87.50 & 86.80 & 86.89 \\
Swin-B \cite{liu2021swin}& 86.8 & 91.85 & 91.74 & 91.62 & 89.84 & 89.01 & 89.07 \\
\midrule
Vim-Ti$^\dag$ \cite{zhu2024vision}& 7.0 & 89.06 & 88.73 & 88.68 & 87.76 & 86.98 & 87.13 \\
VMamba-T \cite{liu2024vmamba}& 30.0 & 93.14 & 92.85 & 92.81 & 91.59 & 90.94 & 91.10 \\
RSMamba-B \cite{chen2024rsmamba}& 6.4 & 94.14 & 93.97 & 93.88 & 92.02 & 91.53 & 91.66 \\
RSMamba-L \cite{chen2024rsmamba}& 16.2 & 95.03 & 94.76 & 94.74 & 92.31 & 91.75 & 91.90 \\
RSMamba-H \cite{chen2024rsmamba}& 33.1 & 95.47 & 95.23 & 95.25 & 92.97 & 92.51 & 92.63 \\
\midrule
DynamicVis-B$^\dag$ & 36.8
& \cellcolor{gray!10}\textbf{96.80} & \cellcolor{gray!10}\textbf{96.66} & \cellcolor{gray!10}\textbf{96.66}
& \cellcolor{gray!10}\textbf{94.41} & \cellcolor{gray!10}\textbf{94.22} & \cellcolor{gray!10}\textbf{94.17}
\\
DynamicVis-B$^\ddag$ & 36.8
& 95.97 & 95.89 & 95.88
& 94.11 & 93.82 & 93.81
\\
DynamicVis-B & 36.8
& \cellcolor{gray!30}\textbf{99.09} & \cellcolor{gray!30}\textbf{99.05} & \cellcolor{gray!30}\textbf{99.05}
& \cellcolor{gray!30}\textbf{96.16} & \cellcolor{gray!30}\textbf{96.00} & \cellcolor{gray!30}\textbf{96.04}
\\
DynamicVis-L$^\dag$ & 91.3
& 96.59 & 96.50 & 96.47
& 94.20 & 93.96 & 93.96
\\
DynamicVis-L$^\ddag$ & 91.3
& 96.34 & 96.20 & 96.16
& 94.08 & 93.82 & 93.79
\\
DynamicVis-L & 91.3
& \cellcolor{gray!70}\textbf{99.12} & \cellcolor{gray!70}\textbf{99.10} & \cellcolor{gray!70}\textbf{99.09}
& \cellcolor{gray!70}\textbf{96.40} & \cellcolor{gray!70}\textbf{96.29} & \cellcolor{gray!70}\textbf{96.28}
\\
\bottomrule
\end{tabular}
\end{table}

\subsubsection{Zero-Shot Image Retrieval}

\textbf{Datasets \& Implementations:}
To validate the generalization capacity and structural integrity of the learned semantic space, we evaluate zero-shot image retrieval on two distinct datasets: BigEarthNet \cite{sumbul2019bigearthnet} (Sentinel-2 multispectral imagery with multi-label land-use annotations) and ForestNet \cite{irvin2020forestnet} (fine-grained, single-label Landsat 8 deforestation imagery). Following standard protocols, the official validation splits serve as queries to retrieve from the test databases. We extract zero-shot features directly from the pre-trained backbones using strictly RGB inputs. Retrieval efficacy is quantified via mAP@20 across three compression formats: 768-dimensional floating-point embeddings, binary embeddings, and 64-bit hash codes (averaging embeddings followed by binarization).

\textbf{Results and Analysis:}
Quantitative results detailed in Table \ref{tab:sota_retrival} reveal that DynamicVis significantly outperforms generic vision models (e.g., ViT \cite{dosovitskiy2020image}) and even surpasses heavily specialized multi-spectral foundation models such as Prithvi \cite{jakubik2023foundation} and SatMAE \cite{cong2022satmae}. This achievement is particularly notable given that DynamicVis achieves this utilizing solely RGB channels, overcoming the inherent limitation of lacking multi-spectral infrared bands. This superiority directly corroborates our core motivation: unlike pixel-reconstruction paradigms (MAE) that force networks to memorize redundant background textures, our MIL pre-training naturally maps imagery into a semantically dense, instance-aware latent space.

\begin{figure}[!tbp]
\centering
\includegraphics[width=\linewidth]{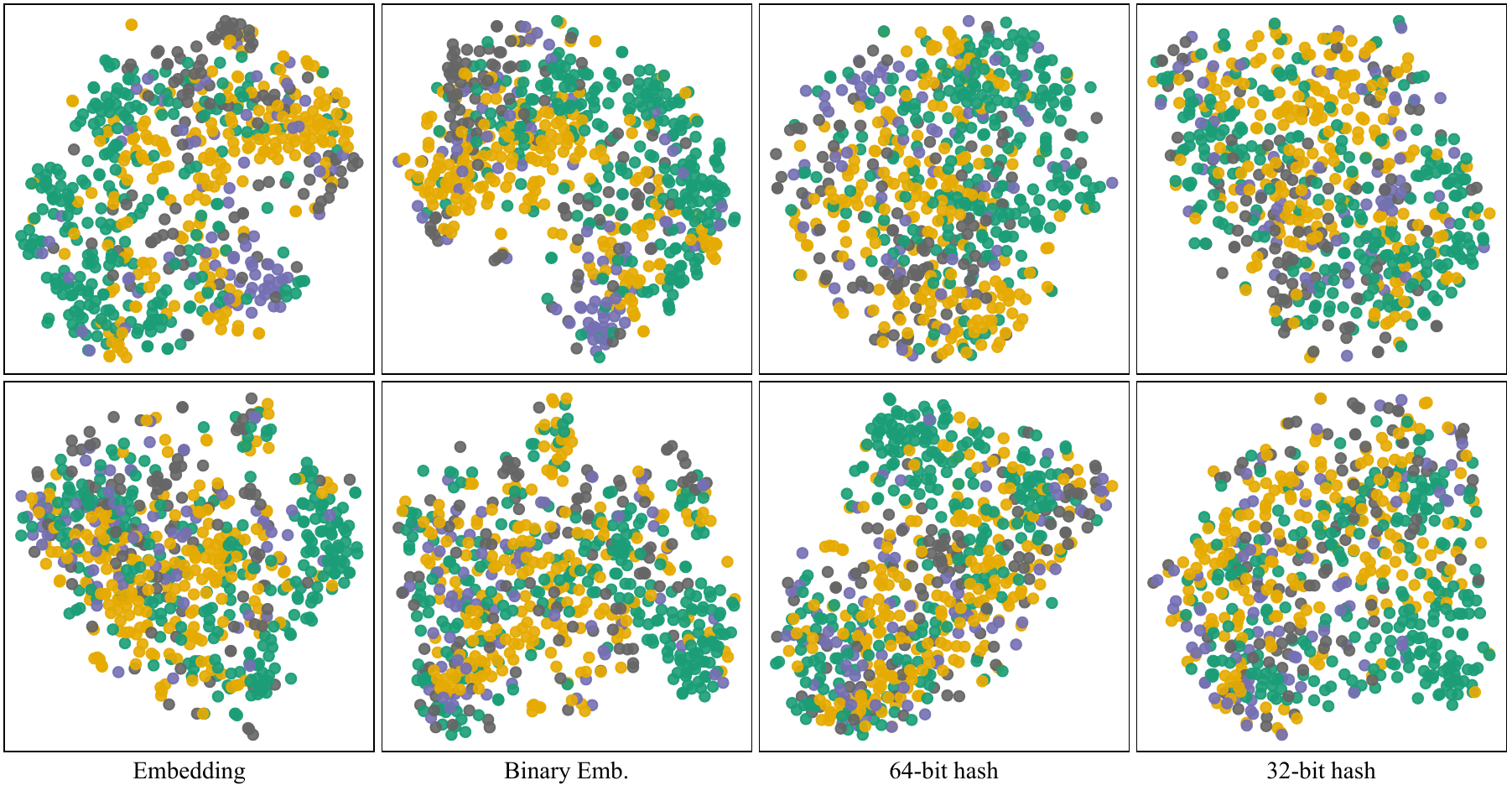}
\caption{t-SNE visualization of zero-shot feature embeddings with ViT and DynamicVis on the ForestNet test set. DynamicVis generates a distinctly more compact and separable latent space compared to ViT, corroborating the efficacy of region-level semantic decoupling.}
\label{fig:vis_retrieval_tsne}
\end{figure}

This structural advantage is vividly illustrated in the t-SNE visualization of the ForestNet test set (Fig. \ref{fig:vis_retrieval_tsne}). Compared to ViTs, DynamicVis produces a highly organized latent distribution exhibiting compact intra-class clustering and decisive inter-class margins. Furthermore, our evaluation of embedding compression confirms that binarization incurs negligible performance degradation while drastically reducing storage constraints. While transitioning to 64-bit trivial hashes induces a minor mAP drop, it demonstrates an optimal trade-off for scaling retrieval across expansive, million-scale Earth observation archives.

\begin{figure*}[!htbp]
\centering
\includegraphics[width=0.93\linewidth]{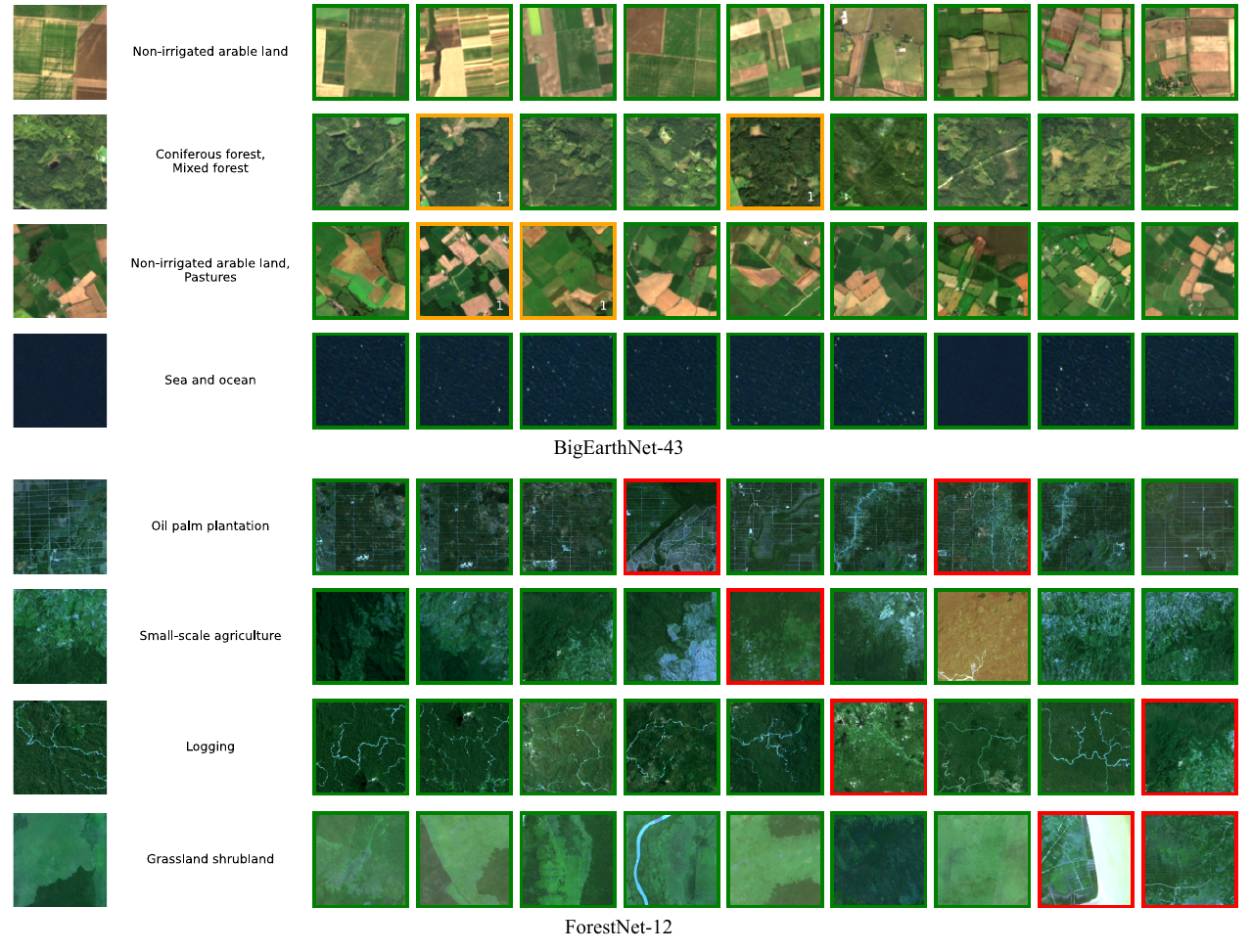}
\caption{Qualitative zero-shot retrieval examples from the ForestNet and BigEarthNet datasets using DynamicVis with 64-bit hash codes. Retrieval quality is delineated by color-coded frames: \textcolor[HTML]{208000}{\textbf{green}} for correct complete matches, \textcolor[HTML]{F9A500}{\textbf{orange}} for partial multi-label matches (numbers indicate intersection count), and \textcolor[HTML]{F60102}{\textbf{red}} for incorrect matches.}
\label{fig:vis_retrieval_results}
\end{figure*}

\textbf{Visualizations:}
Qualitative retrieval results utilizing 64-bit hash codes (Fig. \ref{fig:vis_retrieval_results}) demonstrate the model's practical efficacy. On the multi-label BigEarthNet dataset, DynamicVis routinely retrieves images with near-complete semantic overlap. Conversely, on the fine-grained ForestNet dataset, occasional retrieval errors typically involve non-target images sharing profound textural and spatial similarities with the query (e.g., distinguishing specific types of sparse agricultural plots). These results underscore the inherent challenges of remote sensing retrieval while validating DynamicVis's capability to generalize spatial-semantic priors to unseen domains.

\subsection{Ablation Study and Architecture Analysis}
To systematically validate the efficacy of our proposed mechanisms and justify the specific architectural designs of DynamicVis, we conduct comprehensive ablation studies. Specifically, we isolate the contributions of the dynamic token routing, the SSM scanning strategies, and the Region-Level Meta-Embedding MIL pre-training paradigm. Unless otherwise specified, parametric optimizations are performed on the large-scale AID scene classification dataset to ensure experimental efficiency while providing a robust metric for representation capability.

\subsubsection{Optimization of Dynamic Token Routing}

The Adaptive Token Routing and Incremental Modeling (ATRIM) unit fundamentally dictates the model's capacity to navigate spatial sparsity. We evaluate two critical factors governing its behavior: the stage-wise token reduction ratio and the load-balancing strategy.

\begin{table}[!htpb]
\centering
\caption{Impact of stage-wise token reduction ratios on performance and GPU memory utilization (measured with a $512 \times 512$ input at batch size 1). The default configuration $[7/8, 3/4, 1/2, 0]$ provides the optimal efficiency-accuracy balance.}
\label{tab:ab_reduction_ratio}
\setlength{\tabcolsep}{1.5pt}
\footnotesize
\begin{tabular}{l| c| ccc}
\toprule
Stage-wise Reduction Ratio ($r_1, r_2, r_3, r_4$) & Usage (MB) & P & R & F1 \\
\midrule
$[$0, 0, 0, 0$]$ (No reduction) & 379.99 & 94.41 & 94.22 & 94.17 \\
$[$3/4, 1/2, 0, 0$]$ & 216.58 & 94.32 & 93.95 & 94.05 \\
\rowcolor{gray!15} $[$7/8, 3/4, 1/2, 0$]$ (DynamicVis Default) & 188.71 & 94.11 & 93.82 & 93.81 \\
$[$15/16, 7/8, 3/4, 1/2$]$ & 176.32 & 93.74 & 93.53 & 93.57 \\
$[$31/32, 15/16, 7/8, 3/4$]$ & 174.64 & 92.46 & 92.12 & 92.17 \\
\bottomrule
\end{tabular}
\end{table}

\textbf{Effects of Stage-wise Token Reduction Ratios:} The quantity of selected tokens controls the trade-off between computational scalability and representational integrity. Rather than uniformly discarding tokens across all depths, DynamicVis employs a progressive retention strategy across its four hierarchical stages. As detailed in Table \ref{tab:ab_reduction_ratio}, extreme token reduction ($[31/32, 15/16, 7/8, 3/4]$) minimizes memory usage (174.64 MB) but induces measurable performance degradation. Conversely, retaining all tokens ($[0, 0, 0, 0]$) achieves a marginal F1-score peak (94.17\%) but entirely neutralizes the efficiency advantages, doubling the memory footprint.

Crucially, as previously established, extreme token reduction is tolerable for image-level classification due to its low-information discrimination requirements, but it irreparably fractures the spatial continuum required for dense pixel-level downstream tasks (e.g., semantic segmentation). Therefore, the configuration $[7/8, 3/4, 1/2, 0]$ was selected. This configuration aggressively filters out vast backgrounds (e.g., expansive oceans, bare land) in high-resolution shallow layers, while fully preserving the concentrated, macroscopic semantic structures in the lowest-resolution deep layer, achieving an optimal cross-task balance.

\begin{table}[!t]
\centering
\caption{Evaluation of token loading balance strategies. ``Fixed'' applies constant Gumbel noise, while ``Decay'' progressively reduces the noise scale during training. Standard explicit balancing loss proves suboptimal for handling the severe spatial skewness of remote sensing targets.}
\label{tab:ab_loading_token}
\setlength{\tabcolsep}{8pt}
\footnotesize
\begin{tabular}{l| c| ccc}
\toprule
Strategy & Noise Scale ($\nu$) & P & R & F1 \\
\midrule
Explicit Loss & - & 92.92 & 92.53 & 92.60 \\
Fixed & 0.4 & 92.57 & 92.35 & 92.37 \\
Fixed & 0.2 & 93.42 & 92.72 & 92.89 \\
Fixed & 0.1 & 93.87 & 93.49 & 93.58 \\
Fixed & 0.05 & 93.15 & 92.61 & 92.66 \\
Fixed & 0 & 92.72 & 92.36 & 92.45 \\
\rowcolor{gray!15} Decay (Default) & $0.1 \rightarrow 0$ & 94.11 & 93.82 & 93.81 \\
\bottomrule
\end{tabular}
\end{table}

\textbf{Token Loading Balance Strategy:} The dynamic routing module conceptually mirrors a Mixture-of-Experts (MoE) mechanism. In conventional MoE frameworks, a balanced loading loss is rigidly enforced to prevent expert collapse. However, in Earth observation imagery, semantic information is inherently highly skewed (e.g., a tiny ship may contain 99\% of the semantic value, while the massive ocean contains 1\%). Forcing a mathematically uniform token distribution via an explicit balancing loss disrupts this natural spatial sparsity.

As shown in Table \ref{tab:ab_loading_token}, utilizing a conventional balancing loss yields a suboptimal F1 (92.60\%), reflecting model's inability to concentrate exclusively on sparse targets. Instead, we implement a loss-free strategy by injecting Gumbel-distributed noise into the routing logits. A progressive decay of the noise amplitude ($\nu=0.1 \rightarrow 0$) yields optimal performance (93.81\% F1). This strategy encourages broad exploration in early training phases to discover minuscule targets, before seamlessly transitioning to deterministic exploitation to lock onto highly salient regions during fine-tuning.

\subsubsection{Design of the Region-Aware SSM Scanner}
Recent vision adaptations of Mamba typically mandate exhaustive 4-directional or 8-directional spatial scans to compensate for the non-causal nature of 2D images \cite{zhu2024vision,liu2024vmamba}. We argue that such designs are computationally redundant and conceptually misaligned when combined with adaptive token routing.

\textbf{Scanning Paths and Aggregation:} As shown in Table \ref{tab:ab_scanning_path}, relying solely on a causal \textit{Forward} scan is fundamentally incompatible with spatial data representations. However, adding \textit{Shuffle} paths or complex \textit{Gate} mechanisms (e.g., token-prediction weighted averaging) provides negligible gains. Because our top-$k$ routing mechanism inherently reorganizes the sequence based on continuous semantic importance rather than rigid geometric positions, the selected tokens already transcend traditional 2D grid constraints. Consequently, manual token shuffling becomes redundant. A streamlined \textit{Forward + Reverse} dual-path scan aggregated via simple \textit{Mean} pooling is entirely sufficient, maximizing efficiency without sacrificing contextual modeling capability.

\begin{table}[!tpb]
\centering
\caption{Ablation on SSM scanning paths and aggregation methods. A streamlined dual-path scanning mechanism with mean aggregation is sufficiently robust for dynamically routed token sequences.}
\label{tab:ab_scanning_path}
\setlength{\tabcolsep}{5pt}
\footnotesize
\begin{tabular}{cccc| ccc}
\toprule
Forward & Reverse & Shuffle & Aggregation & P & R & F1 \\
\midrule
\checkmark & & & - & 92.17 & 91.74 & 91.82 \\
\rowcolor{gray!15} \checkmark & \checkmark & & Mean & 94.11 & 93.82 & 93.81 \\
\checkmark & \checkmark & & Gate & 93.83 & 93.69 & 93.65 \\
\checkmark & \checkmark & \checkmark & Mean & 94.12 & 93.91 & 93.96 \\
\checkmark & \checkmark & \checkmark & Gate & 94.14 & 93.84 & 93.90 \\
\bottomrule
\end{tabular}
\end{table}


\begin{table}[!t]
\centering
\caption{Impact of applying dynamic routing across Spatial vs. Channel dimensions.}
\label{tab:ab_sc_scanning}
\setlength{\tabcolsep}{8pt}
\footnotesize
\begin{tabular}{ccc| ccc}
\toprule
Spatial & Channel & Setup & P & R & F1 \\
\midrule
\rowcolor{gray!15} \checkmark & & - & 94.11 & 93.82 & 93.81 \\
& \checkmark & - & 83.31 & 81.69 & 81.91 \\
\checkmark & \checkmark & Serial & 94.52 & 94.25 & 94.29 \\
\checkmark & \checkmark & Parallel & 94.22 & 94.01 & 94.04 \\
\bottomrule
\end{tabular}
\end{table}

\textbf{Spatial vs. Channel Routing:} We explored extending the dynamic selection mechanism to the channel dimension via lightweight MLPs. Table \ref{tab:ab_sc_scanning} reveals that exclusive channel-based routing drastically degrades performance. Fundamentally, remote sensing semantics are spatially localized (e.g., a building occupies specific pixel coordinates, not specific feature channels). Joint Spatial-Channel routing (whether serial or parallel) incurs double the computational overhead for a marginal F1 gain. This confirms our core hypothesis that massive spatial redundancy, rather than channel redundancy, constitutes the primary bottleneck in remote sensing representation learning.

\textbf{Preservation of Global Semantics:} While dynamic routing effectively isolates sparse targets, holistic macroscopic context (e.g., distinguishing a harbor from an inland airport based on surrounding environmental cues) remains imperative. Table \ref{tab:ab_global_semantics} validates that explicitly appending a dense global pooled token ($\mathbf{x}_g$) to the sparse regional sequence ($\mathbf{x}_r$) significantly improves the F1 score from 92.85\% to 93.81\%. Furthermore, the exact insertion position of this global token exhibits minimal variance, reflecting the robust contextual integration capabilities of the bidirectional SSM. The token is appended at the head of the sequence to conceptually align with the highest-ranked regional tokens.

\begin{table}[!tpb]
\centering
\caption{Effects of integrating and positioning the global semantic token during SSM scanning.}
\label{tab:ab_global_semantics}
\setlength{\tabcolsep}{13pt}
\footnotesize
\begin{tabular}{l|ccc}
\toprule
Global Token Position & P & R & F1 \\
\midrule
None (Routing only) & 93.11 & 92.75 & 92.85 \\
\rowcolor{gray!15} Head (Default) & 94.11 & 93.82 & 93.81 \\
Head \& Tail & 94.03 & 93.69 & 93.76 \\
Middle & 94.13 & 93.85 & 93.90 \\
\bottomrule
\end{tabular}
\end{table}

\subsubsection{Decoupling Core Component Contributions}

To explicitly isolate the contribution of each innovation, we trace the performance trajectory from the dense baseline to the fully equipped DynamicVis.

i) \textbf{Baseline ($^\dag$) to Routing ($^\ddag$):} Introducing dynamic token routing dramatically accelerates inference and slashes memory consumption (detailed in Sec. \ref{sec:ab_efficiency}). However, without pre-training semantic priors, the randomly initialized router struggles with zero-shot spatial discrimination, leading to a minor drop in global scene classification (e.g., AID F1 drops from 94.17 to 93.81, see Table \ref{tab:sota_uc_aid}). Yet, for sparse target detection (Table \ref{tab:levirship_sota}), routing immediately boosts small-object $\text{AP50}$ from 78.2 to 80.3. This proves the innate architectural advantage of spatial routing for background noise filtration.

ii) \textbf{Routing ($^\ddag$) to Full Model (MIL Pre-trained):} Once the Region-Level Meta-Embedding MIL paradigm is applied, performance surges across all downstream benchmarks. On AID, the F1 score recovers and leaps from 93.81 to 96.04. On challenging instance segmentation (Table \ref{tab:nwpu_sota}), $\text{AP}_\text{mask}$ climbs from 63.2 to 67.3. This definitively highlights the profound impact of our MIL paradigm: it bridges the semantic gap, explicitly teaching the dynamic router \textbf{how} to decouple foreground instances from redundant backgrounds within the latent space.

\subsubsection{Computational Scalability and Efficiency} \label{sec:ab_efficiency}

A critical claim of DynamicVis is its ability to directly process ultra-high-resolution Earth observation imagery efficiently, a severe bottleneck for current VFMs. Table \ref{tab:ab_backbone_efficiency} compares maximum batch sizes, FLOPs, and throughput on an NVIDIA L20 (48GB) GPU using $512 \times 512$ inputs. While ResNet variants achieve high throughput due to highly optimized parallel convolutions, transformer models (ViT-B) process a mere 86 samples/s with a restricted maximum batch size of 268. In stark contrast, DynamicVis-B achieves 196 samples/s with a massive batch capacity of 998, demonstrating a fundamental structural advantage over dense self-attention mechanisms.

\begin{table}[!tbp]
\centering
\caption{Inference efficiency across diverse architectures with $512 \times 512$ inputs on an NVIDIA L20 (48GB) GPU. DynamicVis establishes a superior balance between model capacity and computational throughput.}
\label{tab:ab_backbone_efficiency}
\setlength{\tabcolsep}{6pt}
\footnotesize
\begin{tabular}{l| cccc}
\toprule
\multirow{2}{*}{Model} & Max & Params. & FLOPs & Throughput\\
& Batch Size & (M) & (G) & (Samples/s)\\
\midrule
ResNet50 \cite{he2016deep}& 642 & 25.56 & 21.47 & 340 \\
ResNet101 \cite{he2016deep}& 608 & 44.55 & 40.92 & 200 \\
ViT-B \cite{dosovitskiy2020image}& 268 & 87.20 & 87.76 & 86 \\
ViT-L \cite{dosovitskiy2020image}& 208 & 305.00 & 362.00 & 26 \\
\midrule
DynamicVis-B$^\dag$ & 186 & 36.76 & 54.28 & 90 \\
\rowcolor{gray!15} DynamicVis-B & \textbf{998} & 36.77 & \textbf{30.07} & \textbf{196} \\
DynamicVis-L$^\dag$ & 132 & 91.27 & 151.00 & 45 \\
\rowcolor{gray!15} DynamicVis-L & \textbf{786} & 91.29 & \textbf{82.31} & \textbf{92} \\
\bottomrule
\end{tabular}
\end{table}

\begin{figure}[!tbp]
\centering
\includegraphics[width=\linewidth]{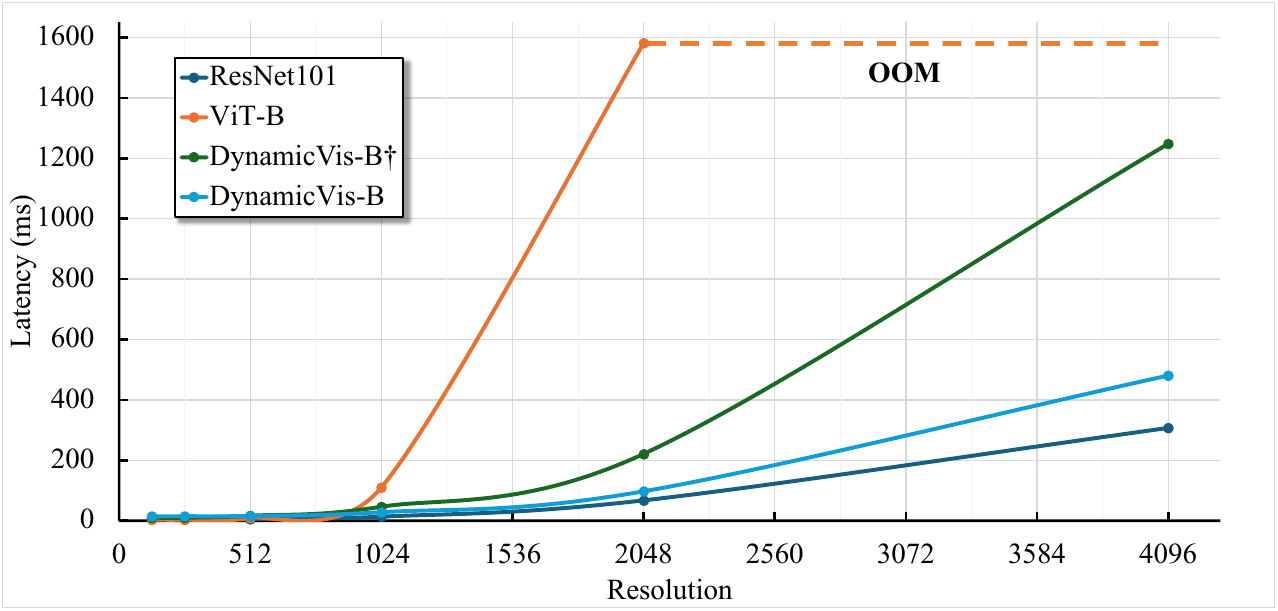}
\caption{Inference Latency (ms) vs. Resolution. ViT exhibits an exponential explosion in latency, whereas DynamicVis preserves strictly linear scalability.}
\label{fig:vis_resolution_efficiency_latency_2}
\end{figure}

\begin{figure}[!tbp]
\includegraphics[width=\linewidth]{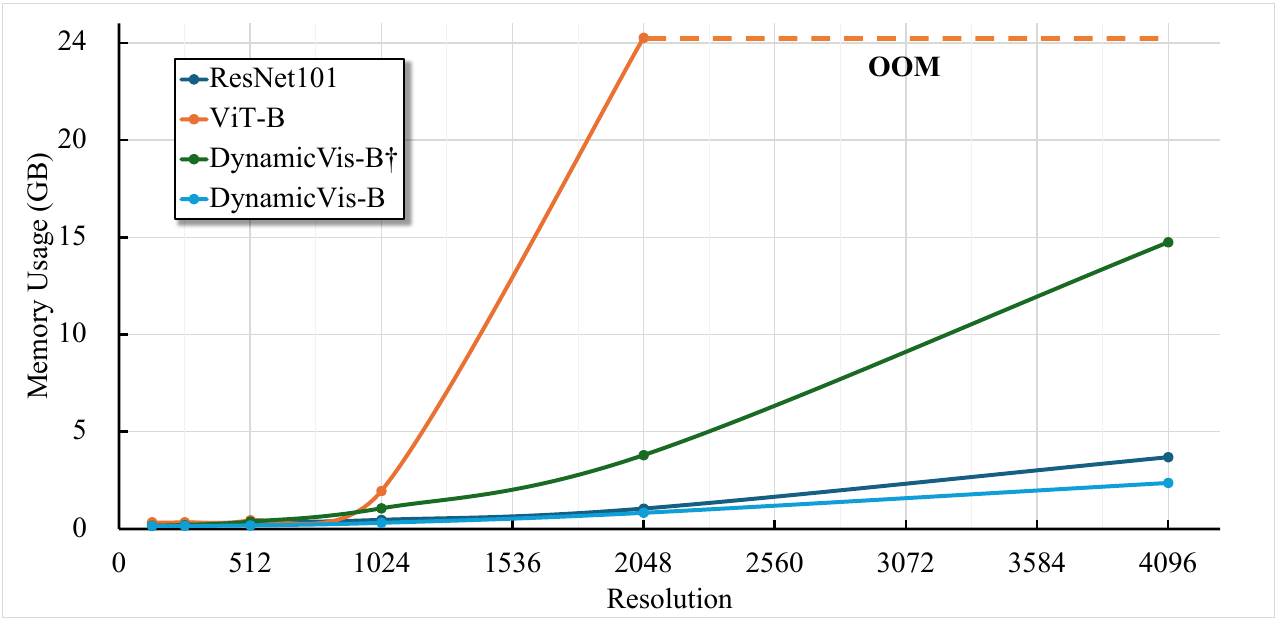}
\caption{GPU Memory Usage (GB) vs. Resolution. DynamicVis drastically undercuts both standard SSMs and ViTs, avoiding memory bottlenecks.}
\label{fig:vis_resolution_efficiency_usage_2}
\end{figure}

More profoundly, we analyze latency and memory scaling against exponentially increasing spatial resolutions (Table \ref{tab:ab_resolution_efficiency}, Figs. \ref{fig:vis_resolution_efficiency_latency_2} and \ref{fig:vis_resolution_efficiency_usage_2}). As resolution approaches $4096 \times 4096$, the $\mathcal{O}(N^2)$ computational complexity of ViT-B results in catastrophic Out-of-Memory (OOM) failures. Conversely, DynamicVis-B scales perfectly linearly, requiring merely 480 ms of latency and 2.3 GB of memory at ultra-high resolutions. Notably, comparing the dense SSM baseline (DynamicVis-B$^\dag$) with the fully routed DynamicVis-B reveals that our dynamic token selection slashes memory consumption at $4096 \times 4096$ by a staggering \textbf{84.4\%} (from 14.7 GB down to 2.3 GB). This compellingly demonstrates that merely transitioning from Transformers to SSMs is insufficient for remote sensing; it is the targeted synergy of linear state-space encoding \textit{and} dynamic spatial routing that unlocks true, unconstrained high-resolution scalability.

\begin{table}[!tbp]
\centering
\caption{Scaling evaluation: Inference latency (ms) and GPU memory usage (MB) across extreme input resolutions. ``OOM'' denotes Out-of-Memory. DynamicVis maintains strictly linear growth, making it highly practical for processing raw, ultra-high-resolution satellite imagery.}
\label{tab:ab_resolution_efficiency}
\setlength{\tabcolsep}{1pt}
\tiny
\begin{tabular}{l| cc| l| ccccc c}
\toprule
Model & Params. (M) & FLOPs (G) & Metric & $128^2$ & $256^2$ & $512^2$ & $1024^2$ & $2048^2$ & $4096^2$ \\
\midrule
\multirow{2}{*}{ResNet101 \cite{he2016deep}} & \multirow{2}{*}{44.55} & \multirow{2}{*}{40.92} & Latency & 5.13 & 5.17 & 5.30 & 13.26 & 67.20 & 307.28 \\
& & & Memory & 300.96 & 309.59 & 328.65 & 470.84 & 1050.65 & 3690.65 \\
\midrule
\multirow{2}{*}{ViT-B \cite{dosovitskiy2020image}} & \multirow{2}{*}{87.20} & \multirow{2}{*}{87.76} & Latency & 3.11 & 3.16 & 9.15 & 109.61 & 1581.18 & OOM \\
& & & Memory & 340.68 & 349.71 & 454.98 & 1958.55 & 25253.80 & OOM \\
\midrule
\multirow{2}{*}{DynamicVis-B$^\dag$} & \multirow{2}{*}{36.76} & \multirow{2}{*}{54.28} & Latency & 9.44 & 9.50 & 16.01 & 45.76 & 220.69 & 1247.37 \\
& & & Memory & 160.51 & 202.57 & 371.98 & 1056.73 & 3793.85 & 14743.85 \\
\midrule
\rowcolor{gray!15}  &  &  & Latency & 14.50 & 14.76 & 15.70 & 27.97 & \textbf{97.18} & \textbf{480.61} \\
\rowcolor{gray!15} \multirow{-2}{*}{DynamicVis-B} &\multirow{-2}{*}{36.77} &\multirow{-2}{*}{30.07} & Memory & 155.10 & 160.07 & 188.71 & 319.25 & \textbf{833.21} & \textbf{2369.06} \\
\bottomrule
\end{tabular}
\end{table}

\subsection{Discussions and Limitations}

By synthesizing linear state-space models with adaptive token routing, DynamicVis circumvents aggressive downsampling heuristics, enabling direct, low-latency ingestion of high-resolution imagery. Furthermore, the Region-Level Meta-Embedding MIL paradigm mitigates the texture-memorization pitfalls of MAE, cultivating a discriminative latent space optimized for sparse instance-level perception.

Despite these advancements, introducing sparse inductive biases involves a natural trade-off regarding topological continuity. While parameter-free residual connections effectively preserve macroscopic global context, delineating highly complex, continuous dense structures (e.g., intricate road networks) only marginally outperforms heavily customized, geometry-constrained models. Because early-stage token routing prioritizes salient discrete instances, it can slightly dilute the high-frequency spatial gradients useful for mapping expansive, non-discrete phenomena. Consequently, peak performance on fully dense prediction tasks may necessitate dynamically lowering token reduction ratios, effectively balancing computational efficiency with structural granularity.

Second, while the MIL pre-training approach is highly data-efficient, its reliance on weak region-level annotations currently limits seamless expansion to web-scale, fully uncurated remote sensing archives. Furthermore, although DynamicVis serves as a robust unified backbone, optimal adaptation across heterogeneous downstream domains still involves task-specific decoders and fine-tuning. Transitioning toward a completely unsupervised, dynamically task-adaptive architecture remains an open research direction.

Looking forward, future work will explore decoupling dynamic spatial perception from supervised semantic priors. Integrating self-supervised modeling exclusively on the \textit{routed} salient tokens, coupled with cross-spatiotemporal contrastive learning, offers a viable trajectory to bypass annotation requirements while retaining strict computational scalability. Additionally, extending the discrete meta-embedding formulation to open-vocabulary region-text alignment will further enhance cross-modal generalization. Ultimately, advancing architectures that seamlessly balance representational capacity, broad generalizability, and low-resource inference remains a pivotal frontier for large-scale Earth observation.

\section{Conclusion}

This paper presents \textbf{DynamicVis}, a highly scalable vision foundation model explicitly designed to navigate the extreme target sparsity and massive spatial redundancy inherent in high-resolution remote sensing imagery. To overcome the computational bottlenecks of uniform dense processing and the semantic ambiguities of pixel-reconstruction paradigms, DynamicVis introduces a Dynamic Region-Aware SSM. Drawing inspiration from the human visual system's selective attention mechanism, this architecture adaptively routes high-salience tokens for computationally intensive state-space refinement, while seamlessly reintegrating unselected background tokens via parameter-free residual connections to preserve macroscopic structural topology. This dynamic routing is empowered by a novel Region-Level Meta-Embedding MIL pre-training paradigm, which leverages a million-scale dataset to explicitly decouple sparse foreground instances from dense backgrounds within the latent semantic space. Extensive evaluations across nine diverse downstream tasks confirm that DynamicVis establishes state-of-the-art performance in multi-granular visual perception, particularly dominating sparse-target scenarios such as small object detection and bi-temporal change analysis. Crucially, the framework achieves unprecedented computational scalability: processing a $2048 \times 2048$ image requires merely 97 ms of latency and 833 MB of memory, accounting for approximately 6\% and 3\% of a ViT-B's footprint, respectively. DynamicVis provides a fundamentally robust and highly practical paradigm for large-scale Earth observation.











\bmhead{Funding}
This work was supported by the National Natural Science Foundation of China (62125102, U24B20177, U25A20401, 624B2017), and Fundamental Research Funds for the Central Universities.

\bmhead{Conflict of interest}
The authors declare that they have no conflict of interest.

\bmhead{Data availability}
All datasets utilized in this research are publicly available and are appropriately referenced in the text.

\bmhead{Code availability}
The code for this study is available at \url{https://github.com/KyanChen/DynamicVis}.

\begingroup
  \small 
  \bibliography{myreferences}
\endgroup

\end{document}